\documentclass[preprint,12pt]{elsarticle}

\usepackage{amssymb}
\usepackage{amsmath,amsfonts}
\usepackage{subcaption}
\usepackage{multirow}
\usepackage{floatrow}
\usepackage{siunitx}
\DeclareMathOperator*{\argmax}{arg\,max}
\DeclareMathOperator*{\argmin}{arg\,min}
\usepackage[T1]{fontenc}

\journal{Computers and Electronics in Agriculture}

\begin{document}

\begin{frontmatter}

\title{Dual-band feature selection for maturity classification of specialty crops by hyperspectral imaging}

\author[inst1]{Usman A. Zahidi}

\affiliation[inst1]{organization={Lincoln Institute of Agri-food Technology, University of Lincoln, Lincoln, UK},
            addressline={Riseholme Park}, 
            city={Lincoln},
            postcode={LN2 2BJ}, 
            country={United Kingdom}}

\author[inst1]{Krystian {\L}ukasik}
\author[inst1]{Grzegorz Cielniak}

\begin{abstract}
The maturity classification of specialty crops such as strawberries and tomatoes is an essential agricultural downstream activity for selective harvesting and quality control (\textbf{QC}) at production and packaging sites. Recent advancements in Deep Learning  (\textbf{DL}) have produced encouraging results in color images for maturity classification applications. However, hyperspectral imaging (\textbf{HSI}) outperforms methods based on color vision as it captures changes in biological attributes, such as an abundance of pigment (anthocyanin and lycopene) and chlorophyll catabolism during the maturity process, by spectral variation. Multivariate analysis methods and Convolutional Neural Networks (\textbf{CNN}) deliver promising results; however, a large amount of input data and the associated preprocessing requirements cause hindrances in practical application. Conventionally, the reflectance intensity in a given electromagnetic spectrum is employed in estimating fruit maturity. We present a feature extraction method to empirically demonstrate that the peak reflectance in subbands such as 500-670 nm (\textbf{pigment band}) and the wavelength of the peak position, and contrarily, the trough reflectance and its corresponding wavelength within 671--790 nm (\textbf{chlorophyll band}) are convenient to compute yet distinctive features for the maturity classification. The proposed feature selection method is beneficial because preprocessing, such as dimensionality reduction, is avoided before every prediction, and the amount of required data for model training is smaller when compared to State-Of-The-Art (\textbf{SOTA}) methods, which eases its deployment. The feature set is designed to capture these traits. The best SOTA methods, among 3D-CNN, 1D-CNN, and SVM, achieve at most 90.0 \% accuracy for strawberries and 92.0 \% for tomatoes on our dataset. Results show that the proposed method outperforms the SOTA as it yields an accuracy above 98.0 \% in strawberry and 96.0 \% in tomato classification. A comparative analysis of the time efficiency of these methods is also conducted, which shows the proposed method performs prediction at 13 Frames Per Second (\textbf{FPS}) compared to the maximum 1.16 FPS attained by the full-spectrum SVM classifier.

\end{abstract}
\end{frontmatter}


\section{Introduction}
\label{sec:sample1}
Strawberry is a high-value and nutritious fruit with formidable economic value. It is non-climacteric, which implies that it is not attributed to post-harvest ripening. Moreover, its shelf life is also limited. Therefore, it is vital to ascertain the right time for harvesting. Conversely, tomatoes are medium-value and climacteric fruits with more significant production and consumption scales.
Moreover, the QC procedures at distributors and supermarkets ensure that anomalous fruits do not end up in the end-user. Thus, maturity stage estimation in both fruits is essential. Due to the sensitivity of the task, expert human harvesters and QC personnel are employed worldwide. However, many key production areas (e.g., UK, US, NL, ES, JP) are now facing severe labor shortages; therefore, developing alternative robotic harvesting and packaging solutions is inevitable. Identifying the correct maturity stage of strawberries and tomatoes is crucial. The maturity classification algorithms have been the topic of interest in the research community for decades \cite{Mehra1969} due to their application in selective harvesting \cite{Yu2019} and quality control procedures \cite{Ashtiani2021}. Many deep learning and feature-based approaches address this problem by employing color and hyperspectral vision.

\subsection{Color Vision}

Contemporary state-of-the-art DL is employed to estimate the maturity stages of specialty crops with feature-based approaches using color images. The application in strawberry maturity classification includes several research contributions such as \cite{Zhou2021} applied YOLOv3 model for classifying eight maturity levels with a mean Average Precision (\textbf{mAP}) of 0.89. \cite{Binder2022} compared feature-based and CNN classification and reported that CNNs better classify unripe strawberries. However, they found that penalized multinomial regression has an accuracy of 86.4 \%. Similarly, \cite{Fitter2019} compared several feature-based approaches with CNN and reported the supremacy of CNNs with 88.0 \% accuracy.\cite{Fan2022} used a dark channel enhancement algorithm to preprocess strawberry images taken at night and finally achieved a ripeness recognition accuracy of over 90.0 \% on YOLOv5. \cite{Jurriaan_Buitenweg2022} assessed maturity classification in ten levels by several pre-trained CNNs and reported EfficientNetB2 as the best classifier with 73.0 \% accuracy. Similarly, \cite{Su2022} applied SE-YOLOv3-MobileNetV1 network to classify tomato maturity with higher speed and mAP of 0.97. \cite{Yu2019} employed Mask-RCNN-based instance segmentation on strawberry with Resnet-50 backbone on 100 images that achieved an average detection precision rate of 95.8 \%, a recall rate of 95.4 \%, and a mean intersection over union (mIoU) rate of 89.8 \%. Similarly, in the domain of tomato classification, many DL models are applied, such as \cite{Khan2023} proposed a convolutional transformer for tomato maturity classification on color images surpassing the state-of-the-art on common benchmark datasets. \cite{Phan2023} employed Yolov5m and Yolov5m combined with ResNet-50, ResNet-101, and EfficientNet-B0, respectively, for classifying tomato fruit on the vine into three classes: ripe, immature, and damaged with an accuracy of 97 \%. \cite{Wang2022} developed a Faster R-CNN model named MatDet for tomato maturity detection, which uses ResNet-50 as the backbone and RoIAlign to obtain more precise bounding boxes and a Path Aggregation Network to address the difficulty of detecting tomato maturity in complex scenarios, their results report mAP of 96.0 \%.\cite{Begum2022} employed VGG, Inception, and ResNet after transfer learning with their dataset for tomato maturity estimation and reported 97.0 \% classification accuracy. Following a feature-based approach \cite{Aguilar2019} applied fuzzy classification architecture on the RGB color model with descriptors to achieve the classification result with $MSE$ of $0.537 \times 10^-3$.
A few classical feature-based classification approaches are also employed, such as \cite{Sherafati2022} utilized $L^*$ and $a^*$ features to estimate six stages of ripening and two stages of storage for their model in TomatoScan, which was also able to determine the ripening stage of tomatoes with an overall accuracy of 75.0 \%. \cite{Goel2015} proposes a Fuzzy Rule-Based Classification approach to estimate the ripeness of tomatoes based on color, which achieved approximately 94.0 \% accuracy in classifying six USDA standardized classes. \cite{ElBendary2015} applied Principal Components Analysis (\textbf{PCA}) in addition to Support Vector Machine (\textbf{SVM}) and Linear Discriminant Analysis algorithms for feature extraction and classification, respectively, and reported 84.0 \% classification accuracy.

\subsection{Hyperspectral Vision}

Although deep learning models produce excellent results with accuracy approaching up to 90.0 \% in RGB images, the annotation process is tedious and time-consuming as strawberries have partial regions of varying maturity stages, which may cause contradictive data in instance classification. Furthermore, researchers typically develop their empirical maturity classes with differing numbers. Therefore, a comparative analysis could not be drawn directly. 

Like computer vision, Deep learning is applied for crop maturity classification in hyperspectral images. For example, \cite{Su2021} developed 1D and 3D residual networks for strawberry hyperspectral data. They reported classification accuracy above 84.0 \% in both models. \cite{Gao2020} first established distinctive wavelengths for strawberry ripeness classification and then applied an AlexNet-based deep learning model on their empirical maturity classes with an accuracy of 98.6 \%. Several alternative approaches, such as SVM, PLS, and PCA, are proven to get promising results for strawberry ripeness classification. \cite{Raj2022} investigated several traits of strawberries, such as water content, solid soluble content, firmness, and ripeness, on data acquired from a set of 43 strawberries. They obtained data through a spectro-radiometer within 300 nm to 2500 nm. Ripeness classification was performed in full-spectrum by SVM and achieved up to 98.0 \% accuracy. \cite{Zhang2016} evaluated strawberry ripeness by HSI systems having focus within two spectral windows of 380 nm-1030 and 874 nm--1734 nm. They defined three classes (ripe, mid-ripe, and unripe), employed PCA for optimal band selection, eventually classified combined windows data by SVM, and reported 85.0 \%  accuracy of their method. \cite{Jiang2016} established multispectral indices to estimate strawberry's maturity. \cite{Shao2020} used PLS and SVM to assess strawberry ripeness with an accuracy of 96.7 \%. In tomato maturity classification \cite{Dai2023} developed a support vector classifier model to determine tomato maturity and demonstrated the classification accuracy using the characteristic wavelength to achieve an accuracy of 95.8 \%. \cite{Jiang2021} developed a semi-supervised algorithm based on Laplacian score and spectral information divergence and the sparse representation model based on class probability for classification with an accuracy of up to 97.0 \%. \cite{Zhao2023} employed random forest, PLS, and recurrent neural networks (\textbf{RNN}) to develop models for predicting the maturity level. Results showed that the RNN model had a classification accuracy of 40.0 \% higher than random forest and 17.0 \% higher than PLS. In the prediction of quality parameters, RNN models had the highest $R^2$ value greater than $0.87$, followed by PLS and random forest models.

A major drawback of hyperspectral imaging-based approaches is the lack of a standard benchmark for evaluation and comparison. Moreover, the reported dataset's sample size is also smaller, making it infeasible for DL approaches to compete with multivariate alternatives.

\subsection{Proposed Approach}

The HSI's rich spectroscopic data helps us understand the biological basis of the fruit maturity process. Numerous biochemical changes occur during fruit ripening. Over 50 polypeptides show prominent changes at different stages of fruit development. Several specific enzymes associated with membranes, anthocyanin synthesis, and sucrose metabolism have been shown to increase in the strawberry during ripening along with chlorophyll decomposition \cite{hancock2020strawberries}. Similarly, lycopene synthesis and chlorophyll catabolism are related to tomato ripening \cite{Su2015}. Levels of both sugars and acids vary significantly in ripe fruit, depending on cultivars and developmental conditions; therefore, these traits are not distinctive for maturity classification. Chlorophyll breakdown is generally an essential catabolic process of leaf senescence and fruit ripening \cite{Hortensteiner2011}. 

The physiological activities during the strawberry and tomato ripening process indicate increased pigments such as anthocyanin and lycopene and decreased chlorophyll with ripening. The pigments have strong reflectance within their bands; therefore, extremum reflectance intensities and their position features imply abundance. We empirically demonstrate reflectance intensities at extremum points in the sub-spectrum of the VNIR range, and their respective positional information in wavelength is essential for classification based on the pigment and chlorophyll abundance. Therefore, we construct a set of statistical features and measure them in varying bandwidths in an iterative feature extraction algorithm that discovers the best feature set and locates their corresponding bands. On the contrary, the SOTA methods rely entirely on reflectance intensities and ignore the extremum reflectance's positional information.

\newpage
The specific contributions of our research are:
\begin{itemize}
    \item  We propose a feature extraction technique that employs the peak reflectance, particularly its position (wavelength) within the pigment band (510 -- 670 nm), to estimate the change in pigment abundance in strawberries. Contrarily, the minimum reflectance and its position within the chlorophyll band (671 -- 790 nm) correlate with the degree of chlorophyll decomposition. Therefore, these features are sufficient to achieve high-quality strawberry maturity classification.

    \item Similarly, for tomatoes, we propose the peak and trough positions within a relatively small pigment band (510 -- 650 nm), peak and trough reflectance together with their positions within a narrower chlorophyll band (651 -- 770 nm) correlate with chlorophyll catabolism. We also show that reflectance data within the visible spectrum, i.e., 450-780 nm, is sufficient for the maturity classification of both fruits. Moreover, it is demonstrated that once the features are selected from the proposed method, any non-linear classifier can achieve equally good results. A comparative analysis of the proposed method with SVM and CNN models concludes that the proposed method is superior to the SOTA in accuracy and Cohen-Kappa (\textbf{$\kappa$}). A comparative time performance analysis of prediction speed in FPS is also performed with SOTA methods.
    
    \item A dataset comprising more than 620 and 540 annotated VNIR (450--850 nm) hyperspectral images of strawberries and tomatoes is made public to enable direct comparisons and benchmarking for further research on this topic. To the best of our knowledge, no public hyperspectral dataset of this size is available for maturity estimation research. The code and data are available from \cite{ZahidiCode2024} and \cite{ZahidiData2024}, respectively.

\end{itemize}

\section{Materials and Methods} \label{sec_sys_arch}

The proposed method selects predefined features from a combination of variable bandwidth in hyperspectral reflectance data. These predefined features comprise statistical measures of the reflectance signature within variable subbands. These features are designed for experimental analysis after understanding the biological phenomena, such as the dependence of the ripening process on pigments and chlorophyll. For benchmarking on our dataset, we also tested SOTA methods based on CNN and SVM classifiers that take full-spectrum voxel data for classification. More than 1100 fruit samples are imaged and annotated by harvesting experts. The subsequent section describes the dataset creation, data analysis, preprocessing, distributions, and similarity analysis. Section \ref{sec_Method} describes the feature extraction method and the model architecture used in this process. Eventually, we include details about the baseline CNN models and their architectures used for comparative analyses.

\subsection{Image Acquisition and Processing}

\begin{figure}[htpb!]
    \centering
\begin{subfigure}[tb!]{0.98\textwidth}
    \centering
    \includegraphics[scale=0.12]{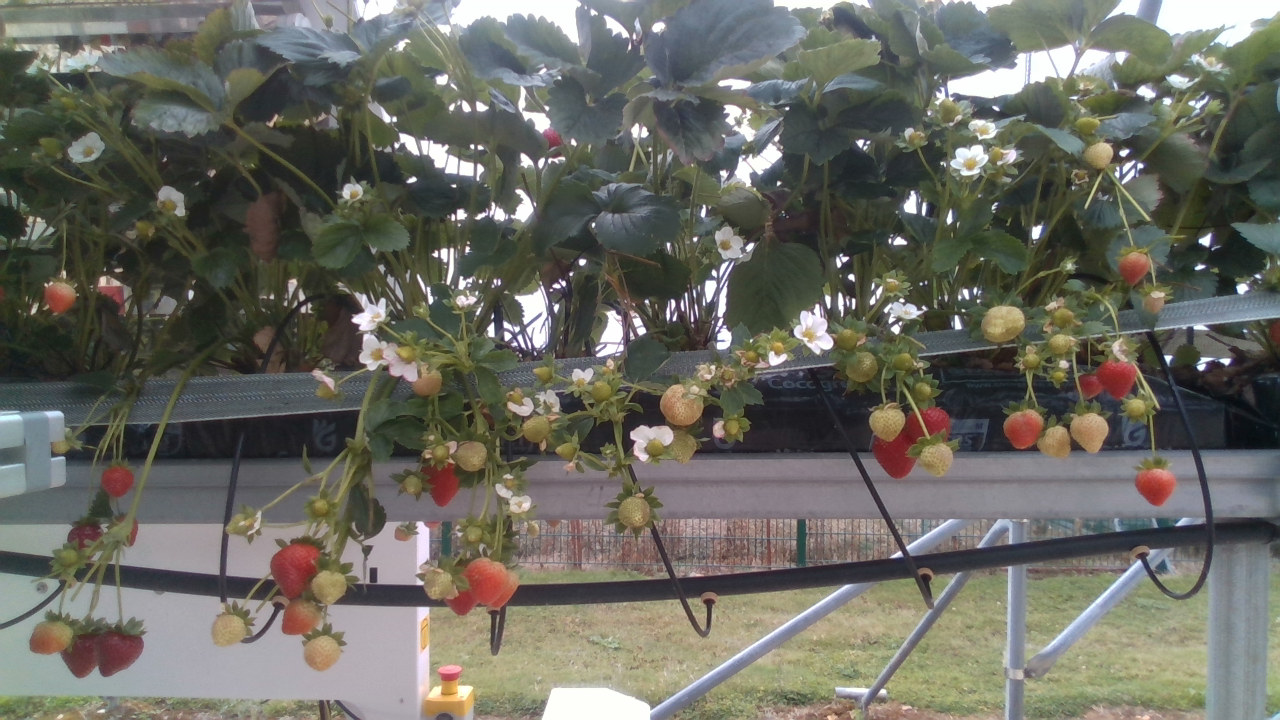}
    \caption{Strawberries grown in a polytunnel at University of Lincoln, UK.}
    \label{fig_avg_f1}
\end{subfigure}
\begin{subfigure}[tb!]{0.49\textwidth}
    \centering
    \includegraphics[scale=0.03]{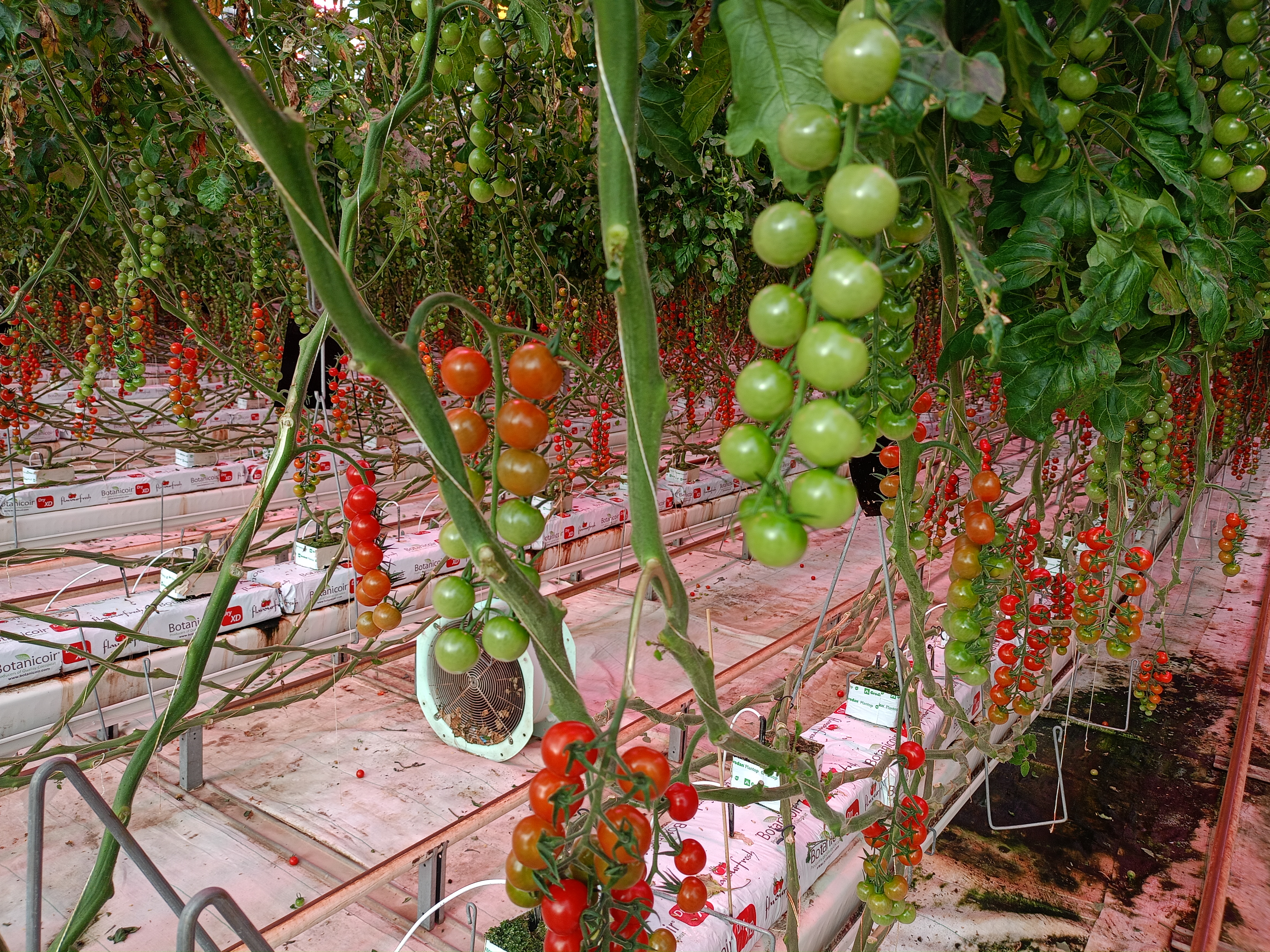}
    \caption{Tomatoes in Glasshouse (FlavourFresh, UK)}
\end{subfigure}
\begin{subfigure}[tb!]{0.49\textwidth}
    \centering
    \includegraphics[scale=0.15]{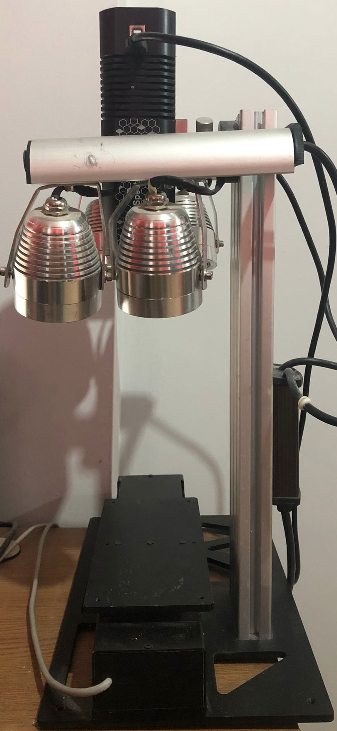}
    \caption{Line-scan hyperspectral camera.}
\end{subfigure}
\caption{The image acquisition camera, the University of Lincoln and FlavourFresh Salads growing facilities} 
\label{fig_hsi_imager}
\end{figure}

The images were taken by a Visible and Near-Infrared line-scan hyperspectral imaging system shown in Fig.~\ref{fig_hsi_imager} (c). This system consisted of a spectrograph from LGL AB, Sweden, a charge-coupled device (CCD) camera from Basler, four halogen lamps, a linear actuator, and a computer interface. The camera has a slit sampling of 368 pixels with a spectral resolution of 2.5 nm in dual binning. We constructed images by scanning 120 lines so the image has a spatial resolution of $120\times368$ px and a spectral resolution of 400 bands between 450 and 850 nm. They were spectrally cropped to avoid sensor noise due to lower spectral response at band extreme edges. The exposure time was set to 150 ms, and the distance between the lens and the translating platform was 22 cm. The non-uniform luminance distribution and dark 

\noindent current from the camera and the hyperspectral images required calibration before spectral reflectance extraction. The raw images were corrected using 

\begin{equation}
R=\frac{I_{raw}-I_d}{I_w-I_d}
\label{eq_calibration}
\end{equation}

\noindent Eq.~ \ref{eq_calibration}, where $R$ is spectral reflectance, $I_{raw}$ is the raw image intensity, while $I_w$ and $I_d$ are white and dark references, respectively. The light scattering correction was applied using Multiplicative Scattering Correction (MSC) \cite{Martens2003}, and the Savitzky–Golay filter smoothed the reflectance.

The imaging system is connected to a personal computer with an Intel(R) Core(TM) i7-9700K CPU @ 3.60GHz, an eight-core machine equipped with 64 GB memory, and an Nvidia RTX-4090 GPU with 12 GB GDDR memory. The system runs Ubuntu 22.04 OS and hosts Nvidia CUDA 11.4. The model prediction FPS is measured on this machine for all models.

\subsection{Dataset Preparation and Analysis}

\begin{figure}[htpb!]
\centering
\begin{subfigure}[tb!]{0.49\textwidth}
    \centering
    \includegraphics[scale=0.25]{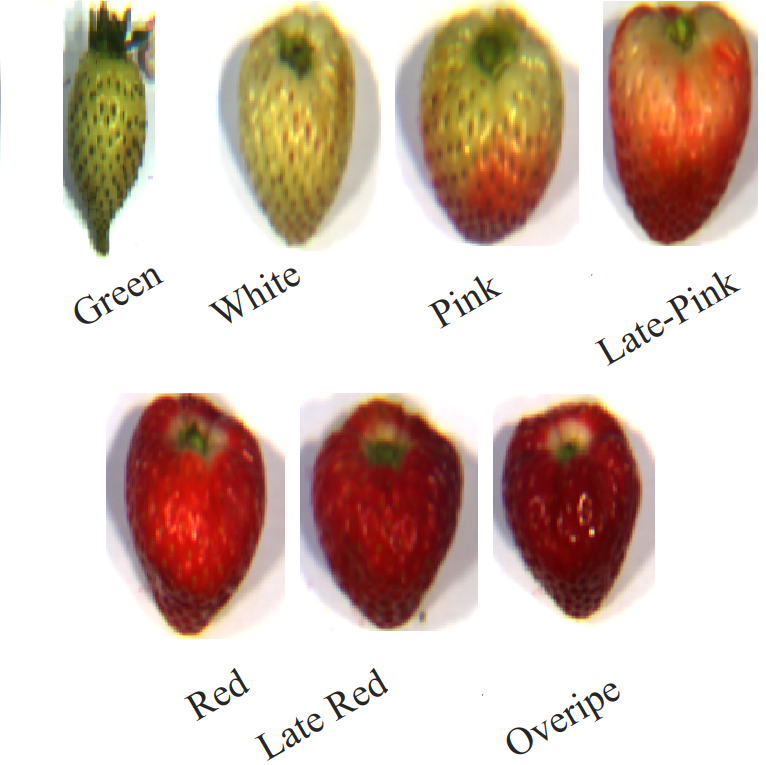}
\end{subfigure}
\begin{subfigure}[tb!]{0.49\textwidth}
    \centering
    \includegraphics[scale=0.235]{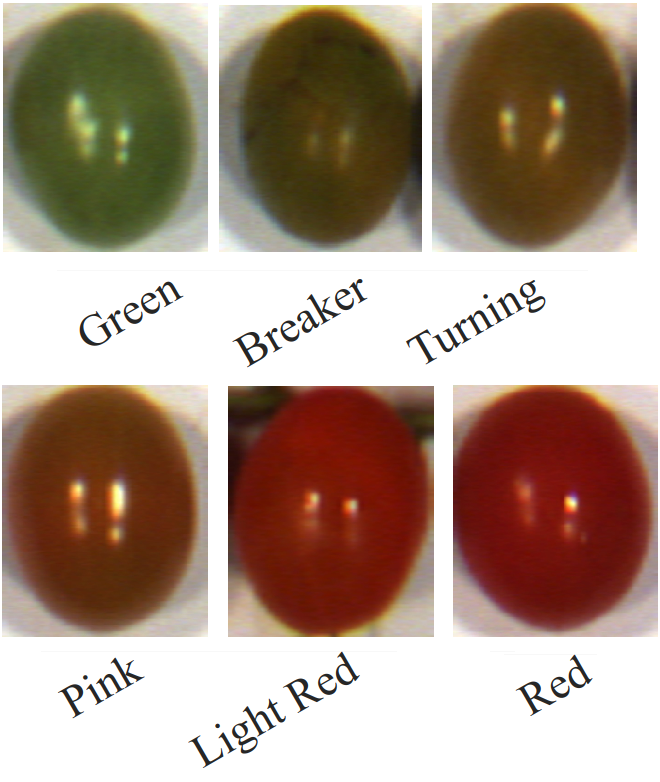}
\end{subfigure}

\begin{subfigure}[tb!]{0.49\textwidth}
    \centering
    \includegraphics[scale=0.39]{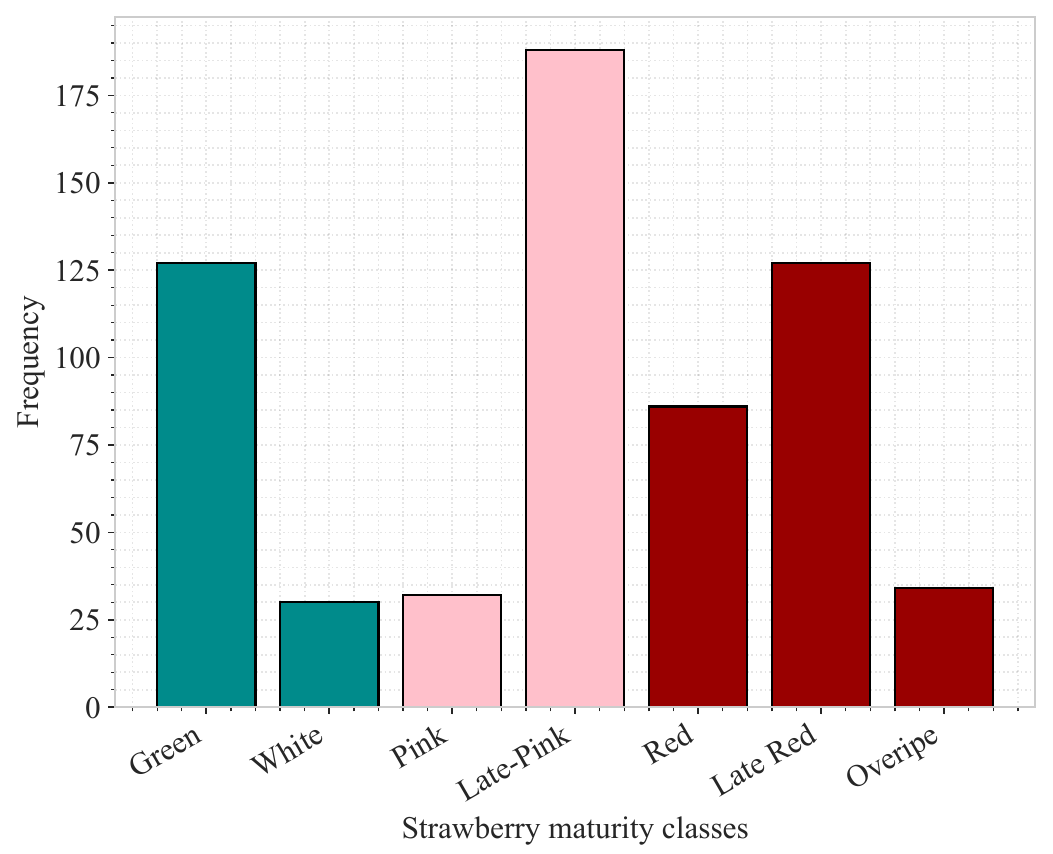}
\end{subfigure}
\begin{subfigure}[tb!]{0.49\textwidth}
    \centering
    \includegraphics[scale=0.39]{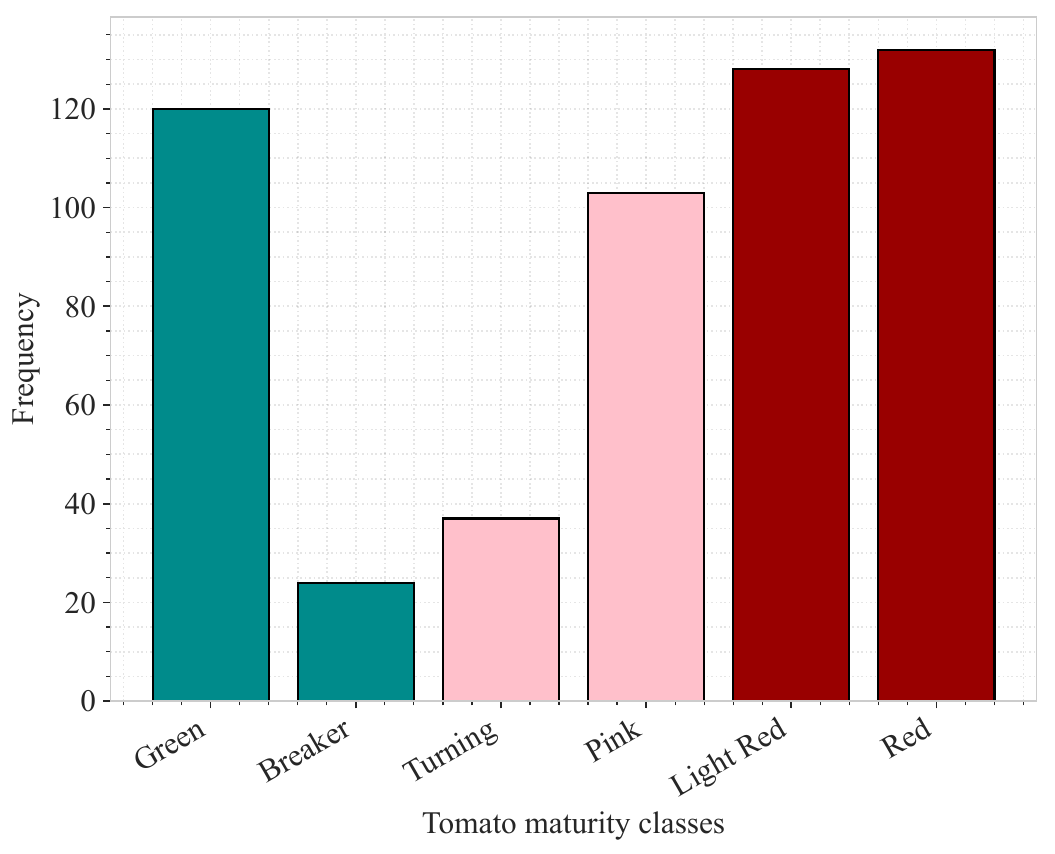}
\end{subfigure}
\caption{Class-wise instance images and dataset distribution in strawberry (left) and tomato (right).} 
\label{fig_tom_straw_dist}
\end{figure}

There are no official standards for the classification of the various maturity stages of strawberries. Therefore, it varies from grower to grower; furthermore, supermarkets follow their own QC specifications for identifying unripe fruits. The traditional method of judging strawberry maturity stages manually determines strawberries' appearance, color, texture, flavor, and firmness, which is time-consuming \cite{Zhou2021}. Based on expert harvesters' opinions, we define seven maturity classes for strawberries ranging from \emph{Green} to \emph{Overripe}. The details of individual classes, their hyperspectral signatures, example class instance images, and their distribution in our dataset are given in Fig.~\ref{fig_tom_straw_dist}. Our dataset comprises 624 Driscoll's strawberries of Katrina and 

\begin{table}[!htpb]
\centering
\caption{Summary of the dataset size. A single strawberry sample is present in an image with a resolution of $120\times368\times400$. Multiple tomatoes are in the same image; therefore, a single instance is cropped from it.}
\label{tab_dataset_summary}
\scalebox{0.75}[0.75]{
\begin{tabular}{|c|c|c|c|} 
\hline
\multirow{2}{*}{Dataset}&\multicolumn{2}{c}{Image \#}\vline &\multirow{2}{*}{Instances}\\ 
\cline { 2 - 3 } & train & test &\\
\hline Strawberry (Driscoll's Katrina) & 247 & 124& $371$  \\ 
\hline Strawberry (Driscoll's Zara) & 169 & 84& $253$  \\ 
\hline Tomato (Piccolo) & 364 & 180 & $544$ \\ \hline
\end{tabular}}\\
\label{tab_fruit_dataset}
\end{table}

\noindent Zara types, where samples were harvested during fall 2021 and summer 2023, which were grown in the research strawberry farms at the University of Lincoln as shown in Fig.~\ref{fig_hsi_imager} (a). The types and distribution of our dataset are given in Tab.~\ref{tab_fruit_dataset}. Contrarily, the United States Department of Agriculture (\textbf{USDA}) officially established maturity classification standards for tomatoes in \cite{USDA1997}. Following this standard, there are six classes, from \emph{Green} to \emph{Red}, in our dataset. The tomato dataset comprises 544 images. The ground truth (GT) of maturity classes for both strawberry and tomato samples was annotated manually by the expert harvesters on an instance basis. GT construction of the tomato images was performed according to the USDA specifications described in the document to ensure conformity. Tomatoes were sourced from Glasshouse at FlavourFresh Salad growers, as shown in Fig.~\ref{fig_hsi_imager} (b). The images were taken in a lab environment, and the background and leaves were removed by adaptive thresholding of Normalized Difference Vegetation Index \textbf{NDVI} \cite{Huang2021}. 

\begin{figure}[htpb!]
    \centering
\begin{subfigure}[tb!]{0.49\textwidth}
    \centering
    \includegraphics[scale=0.442]{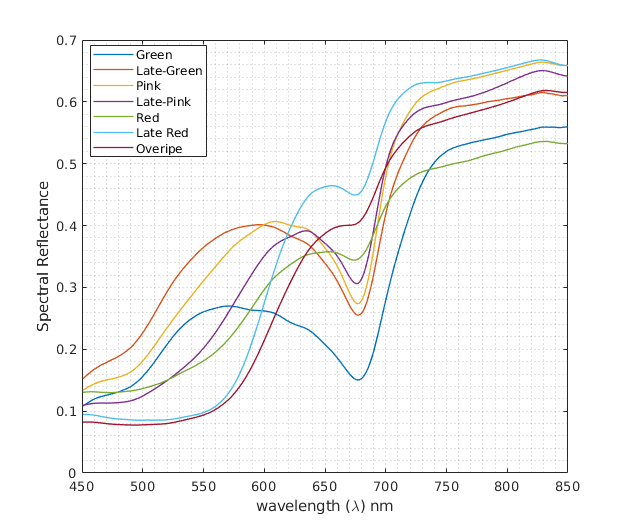}
    \caption{Strawberry}
\end{subfigure}
\begin{subfigure}[tb!]{0.49\textwidth}
    \centering
    \includegraphics[scale=0.442]{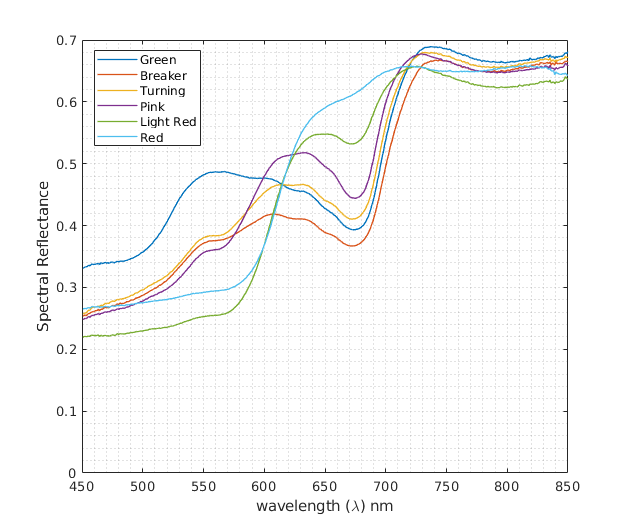}
    \caption{Tomato}
\end{subfigure}
\caption{The Ground Truth spectral signature for individual classes. The spectra show a shift in peak from left to right between 510 and 670 nm with maturity. Furthermore, a rise in trough reflectance due to chlorophyll decomposition is also observed between 670 and 790 nm in strawberries and tomatoes.} 
\label{fig_straw_tom_dist}
\end{figure}

\begin{figure}[htpb!]
    \centering
    \includegraphics[scale=0.16]{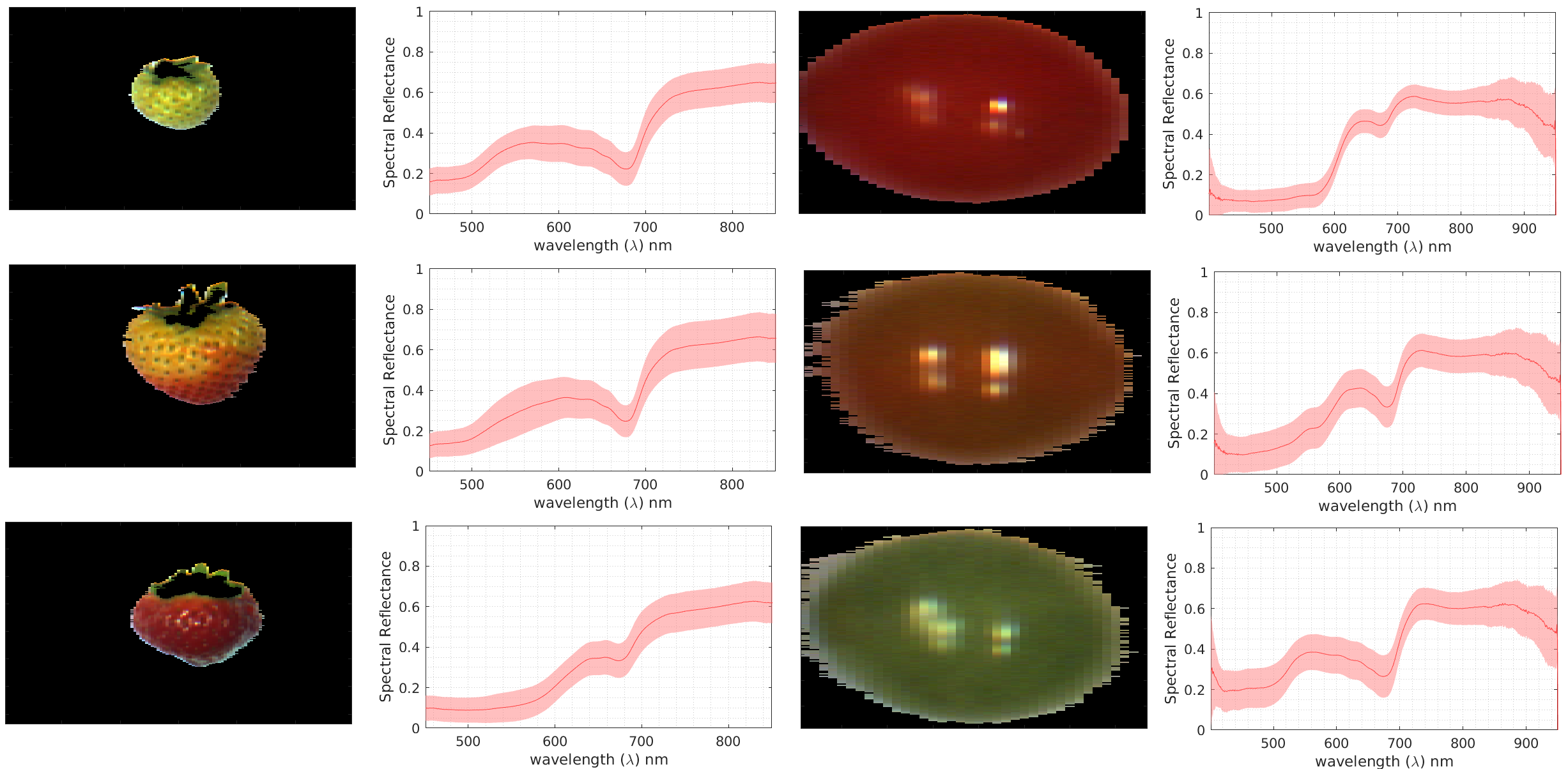}
    \caption{strawberry (column 1) and tomato (column 3) RGB images constructed from HSI after background and leave regions subtracted from the original image. Strawberry (column 2) and tomato (column 4) mean reflectance and standard deviation of segmented strawberry instance spectra. The degree of variance across the VNIR spectrum within three different classes is shown.}
    \label{fig_straw_refl}
\end{figure}

The mean spectra for the fruit region in images are labeled as individual fruit's GT spectra after removing the leaves, as shown in Fig.~\ref{fig_straw_refl}. The mean reflectance of all the fruit spectra belonging to a particular class is called class GT spectra, as shown in Fig.~\ref{fig_straw_tom_dist}. These class spectra show a pattern of peak shift in pigment bands and reflectance hike in the chlorophyll band of both fruits. The number of both fruit samples is cumulatively above 1100.

\begin{figure}[htpb!]
    \centering
\begin{subfigure}[t!]{0.49\textwidth}
    \centering
    \includegraphics[scale=0.38]{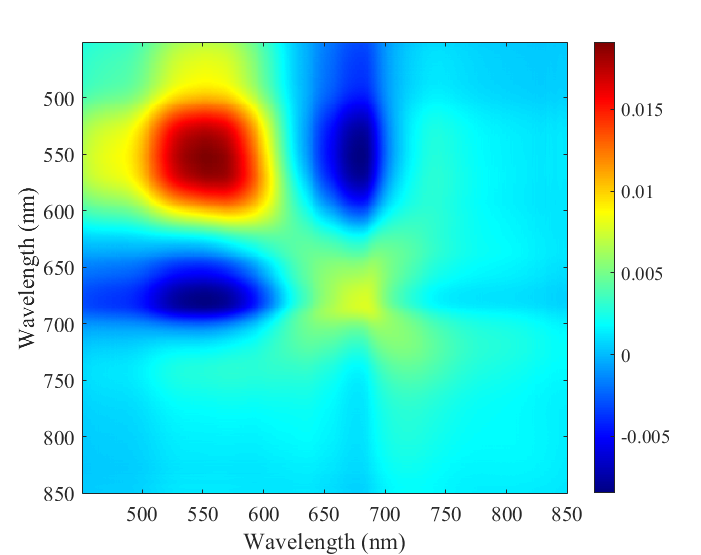}
\end{subfigure}
\begin{subfigure}[t!]{0.49\textwidth}
    \centering
    \includegraphics[scale=0.38]{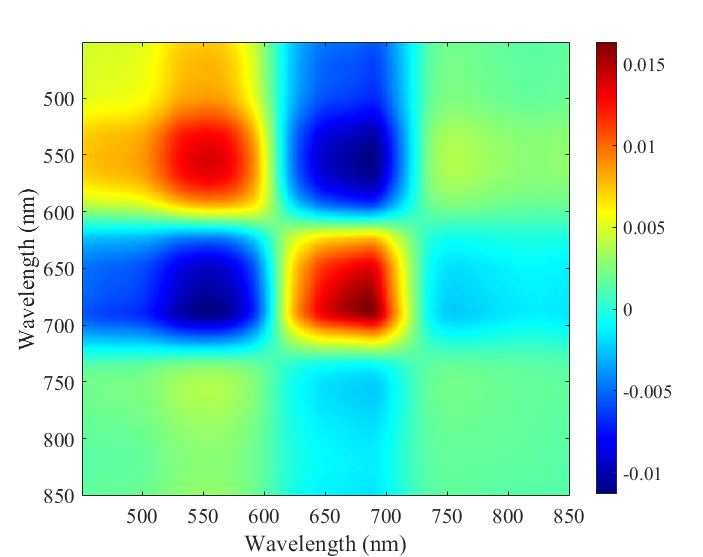}
\end{subfigure}
\caption{A Band-to-band covariance of strawberry (left) and tomato (right) show that similar bands have higher and lower value regions between 520 and 600 nm in both datasets. The tomato dataset also shows high covariance between 630 and 730 nm, which is relatively lower in strawberries. Lower covariance between 450--630 nm and 620--720 nm implies potential mutual exclusion. } 
\label{fig_covaraince}
\end{figure}

In the strawberry dataset, the instances for all classes have a decent sample size; however, the \emph{White}, \emph{Pink}, and \emph{Overripe} are relatively lower. Similarly, for the tomato dataset, the frequency is relatively low in the \emph{Breaker} and \emph{Turning} classes, as shown in Fig.~\ref{fig_tom_straw_dist}. The band-to-band covariance of hyperspectral data highlights the hotspot band regions of higher information, shown in Fig.~\ref{fig_covaraince} for instance-level spectra of strawberries and tomatoes. The description of the band covariance method in HSI can be found at \cite{ArcMap2024}. The value of covariance is similar in both fruits. The relative distribution of band covariance shows some similarity in bands where higher covariance appears between 520 and 600 nm. The tomato dataset also shows high covariance between 630 and 730 nm, which is relatively lower in strawberries. Both fruits are similar in color; however, there are a few differences, such as the strawberry has more textural information due to variable maturity regions, the presence of achenes, and a relatively diffuse Bi-directional 

 \noindent Reflection Distribution Function (\textbf{BRDF}), \cite{Montes2012}. The tomato possesses a relatively specular BRDF and has smoother and relatively uniform color distribution on its skin. The clustering of higher covariance in two regions gives clues about the changes in pigment and chlorophyll in dataset samples.
 
\subsection{Method} \label{sec_Method}

The essential characteristic of the proposed method is feature selection and extraction within various reflectance subbands. Ideally, the hyperspectral unmixing of pigment, chlorophyll endmembers, and their abundance map by employing \cite{Nascimento2005},\cite{Burkni2018}, and \cite{Heinz99} could give the required abundance information for maturity classification. However, this is computationally expensive and practically infeasible for deployment. Our feature extraction process finds minimal features correlated with the pigment and chlorophyll abundance, making it suitable for deployment. 

\subsubsection{Feature Extraction} \label{sec_feature_extraction}

A hyperspectral image $\mathcal{I} \in \mathbb{R}^{H\times W\times B}$ consists of  $d=H{\times}W$ spectra $\textbf{s} \in \mathbb{R}^{d \times B}$ represented as vectors of length $B$, where $H$,$W$ and $B$ represent height, width and bands, respectively. A physical scene renders its texture due to the variability of constituent materials called endmembers \textbf{e}. Linear Mixing Model (LMM) represents an image by pixel-wise spatial distribution of abundances $\textbf{a}$ corresponding to the spectral set of $\textbf{e}$. Image pixel reflectance $s_i$ in LMM is given in Eq.~\ref{eq_avg_rec}, where $n$ represents additive 

\begin{equation}
s_i=\sum_{i=1}^{h{\times}w} a_i e_j + n
\label{eq_avg_rec}
\end{equation}

\begin{figure}[htpb!]
    \centering
\begin{subfigure}[tb!]{\textwidth}
    \centering
        \includegraphics[scale=0.26]{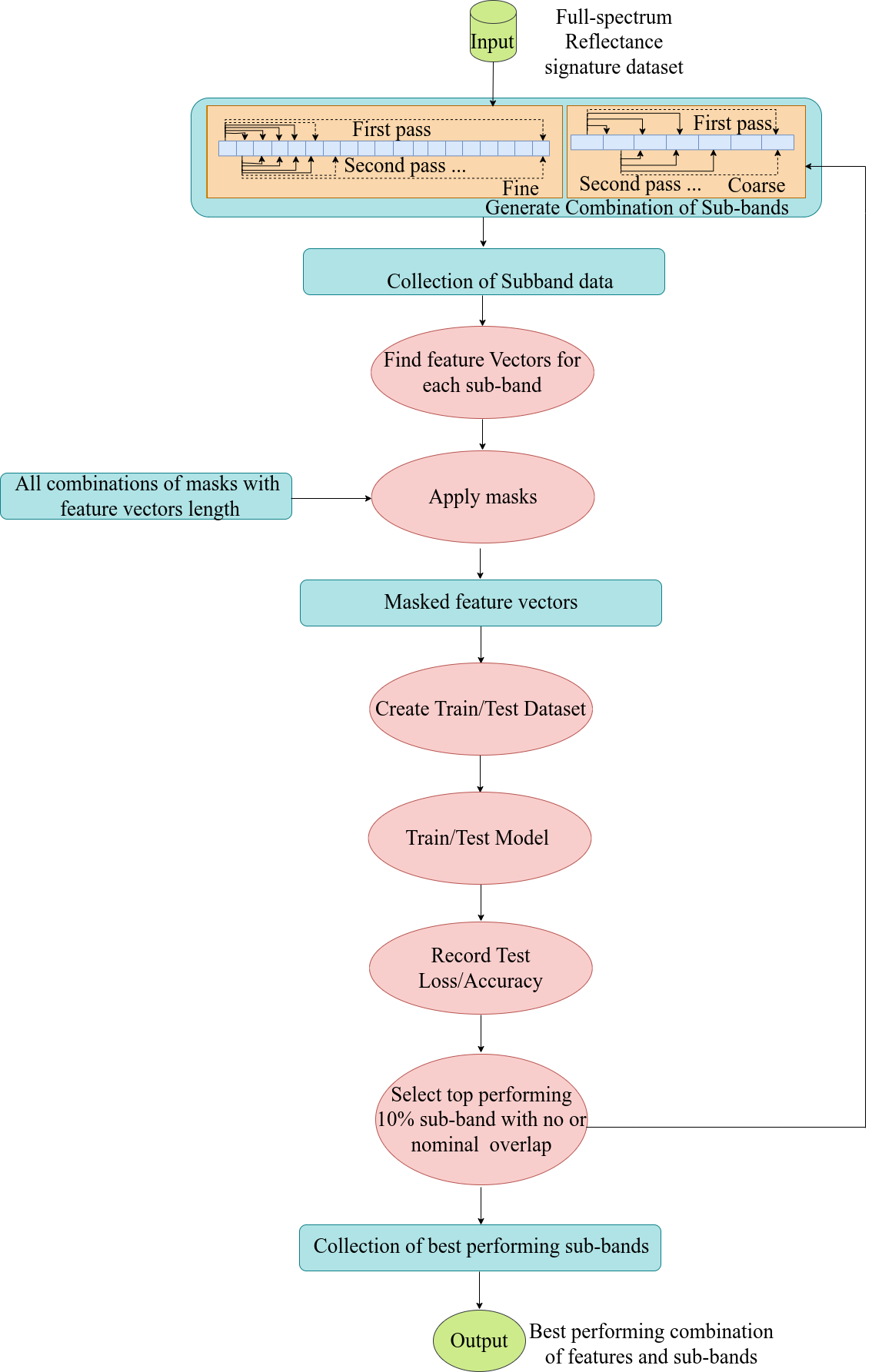}
        \label{fig_feature_flow}
\end{subfigure}

\begin{subfigure}[tb!]{\textwidth}
    \centering
    \includegraphics[scale=0.38]{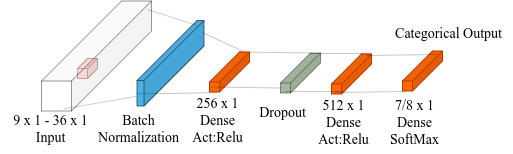}
    \label{fig_fcn_arch}
\end{subfigure}
\caption{The automated feature selection procedure overview (top), model employed in the feature search (bottom).} 
\label{fig_feature_overview}
\end{figure}

\noindent white noise, $i$ represents each voxel of the HSI image, and $j$ represents the number of endmember spectra. The correlation of fruit maturity $M_F$ with pigments such as anthocyanin and lycopene abundance $a_{p}$ and chlorophyll abundance $a_{c}$ is established in biological literature, \cite{hancock2020strawberries} and \cite{Su2015}, as referred earlier. Hyperspectral unmixing could be applied to extract these abundances for corresponding pigment and chlorophyll endmembers. However, in our case, it requires full VNIR spectral resolution between 450 and 850 nm. We demonstrate that the abundances $a_p$ and $a_c$ correlate with fewer statistical measures within bands $b$ such that $b \subseteq B$. Let $R_F$ be the spectral reflectance of fruit, then the feature vector $\mathcal{F}_M$ is defined in Eq.~\ref{eq_feature_vec}. The feature vector comprises the reflectance extremum points to capture the peak and trough reflectances, argmin, and argmax for determining the wavelengths of extremum, which is given in $\mu m$ to keep the value between 0 and 1. The statistical measures are mean, median, and area under  $R_F$ within band range $B$, including the higher order statistics, skewness, and kurtosis.

\begin{equation}
\begin{split}
\mathcal{F}_M(R_F,B)=[max(R_F),min(R_F),\argmax(R_F),\argmin(R_F),mean(R_F),\\ 
median(R_F), area(R_F),skewness(R_F),kurtosis(R_F)]^T
\end{split}
\label{eq_feature_vec}
\end{equation}

\begin{figure}[tb!]
    \centering
    \includegraphics[scale=0.45]{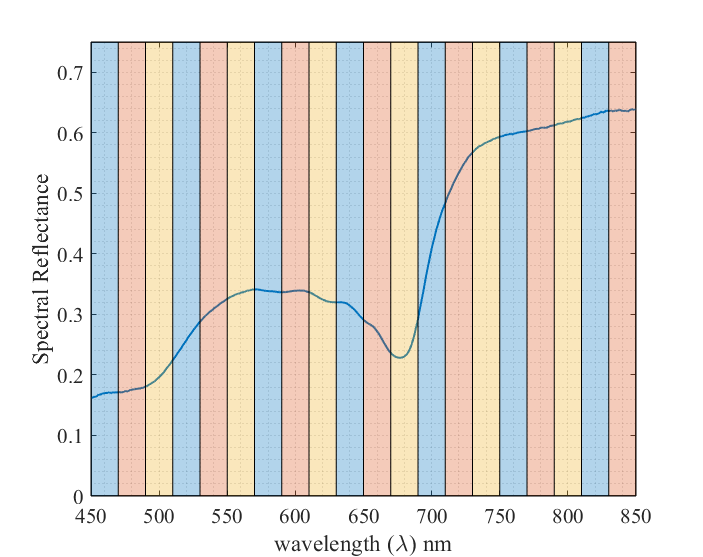}
    \caption{The reflectance between 450--850 nm is divided into 20 subbands of fixed 20 nm width. In every iteration of the feature extraction method, the bandwidth is increased by incorporating the consecutive band.}
        \label{fig_fruit_bands}
\end{figure}

\noindent In our experiments, $B=400$ {nm} is divided into twenty equal bands, each with a width of $b_w=20$~nm, as shown in Fig.~\ref{fig_fruit_bands}. The selection of fixed bandwidth is based on a trade-off between accuracy and computational cost; the larger the bandwidth, the smaller the time it consumes during the search. However, the smaller bandwidth would give a more precise window size, and larger ones may include some unnecessary wavelengths. We define $b^L$ as

\noindent a variable length successive band add-ons such that the superscript $L=n \times b_w$ where, $1 \leq n \leq 20 $. We also define a binary mask $\mathcal{M}$ as all possible combinations of the length of $\mathcal{F}_M$ and $D$ is a collection of $d$, i.e., $D=<d>$, so $D_L$ is the collection of variable datasets for fruit spectra for bands $b^L$ as defined in Eq.~\ref{eq_dataset}. 

\begin{equation}
D_L(\mathcal{M},R_F,b^L)=\mathcal{M} \times \mathcal{F}_M(R_F,b^L)
\label{eq_dataset}
\end{equation}

We construct a fully connected neural network (FCN) whose architecture is shown in Fig.~\ref{fig_feature_overview} (b). This network has the categorical cross-entropy loss function $\mathcal{L}$ for $k(M_F)$ or conventionally $k$ classes, as given in Eq.~\ref{eq_loss_func}, where $t_i$ is the ground truth label and $p_i$ is the Softmax probability for the $i^{\text {th }}$ class, where the input vector of $p$ belongs to $D_L$.

\begin{equation}
\mathcal{L}(D_L)=-\sum_{i=1}^k t_i \log \left(p_i\right) \text {, for } \mathrm{k} \text { classes, }
\label{eq_loss_func}
\end{equation}

The feature extraction method seeks the best subset of features given in Eq.~\ref{eq_feature_vec} and the subband ranges $b^L$. The feature search is implemented by applying all possible combination masks denoted as $\mathcal{M}$. The search starts from the first band and increments the width by $b_w$ in each iteration such that the bandwidth is $b^L_i$ for $\text{i}_{th}$ iteration until it reaches the maximum band of the spectrum, as shown in Fig.~\ref{fig_feature_overview} (a). The spectral dimension of dataset $d$ is sliced according to the $b^L_i$, and the feature vector  $\mathcal{F}_{Mi}$ of the iteration is computed and multiplied with the iteration mask $\mathcal{M}_i$ before it is passed to the FCN model for classification and evaluation. A list of the iteration attributes, such as bandwidth, mask, test accuracy, and loss, is maintained as shown in Fig.~\ref{fig_feature_overview}. An exhaustive search is performed, and the cost function is minimized for all the datasets belonging to $D_L$; the feature set with maximum accuracy is selected, and its list is sorted in ascending order. Top q ($default=10$), features are selected from the list as per user configuration. The second pass picks all mutually exclusive or nominally overlapping in terms of $b^L$ and the highest accuracy. Finally, datasets with minimal $\mathcal{L}(D_L)$ are chosen from the list, and their mask's nonzero features are concatenated before model training in the next step. An example of two-band concatenation and their respective masks and accuracies are illustrated by

\begin{figure}[htbp!]
    \centering
\begin{subfigure}[tb!]{0.49\textwidth}
    \centering
        \includegraphics[scale=0.39]{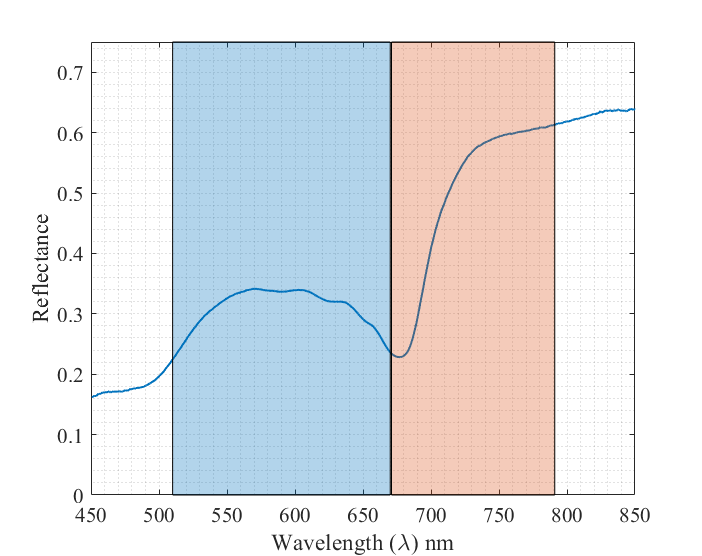}
        \caption{Strawberry.}
        \label{fig_straw_two_bands}
\end{subfigure}
\begin{subfigure}[tb!]{0.49\textwidth}
    \centering
        \includegraphics[scale=0.39]{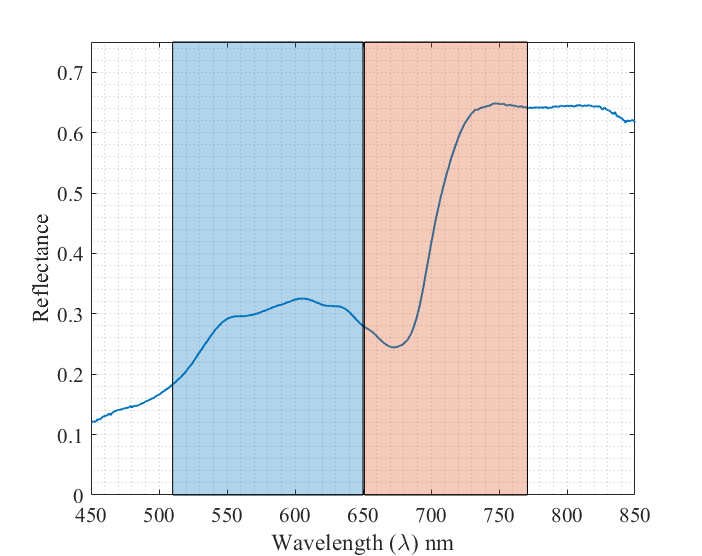}
        \caption{Tomato.}
        \label{fig_tom_two_bands}
\end{subfigure}
\caption{Two prominent bands with the highest accuracy are depicted for unripe (\emph{Green}) strawberry and tomato instances. The Pigment band is shown in blue, while the Chlorophyll band is shown in light orange. The width of both these bands is relatively smaller in tomatoes.}
\label{fig_feature_concat}
\end{figure}

\noindent example in Fig.~\ref{fig_feature_concat}. It is shown that the pigment band only employs two elements of the feature vector $\mathcal{F}_{M}$, namely the minimum reflectance of the band and the wavelength of the maximum reflectance. The chlorophyll band, however, requires four features at most: maximum, minimum reflectance, peak, and trough wavelength position. 

Fig.~\ref{fig_feature_overview} gives the feature extraction search flow; the search algorithm takes the full-spectrum data, ranging between 450 and 850 nm. The fixed width $b_w=20$ nm, so the spectrum has 20 fixed subbands. The search starts by selecting the first (left-most) bands as shown in the entity "Generate Combination of subbands" in Fig.~\ref{fig_feature_overview} (a). The feature vector of reflectance in this subband is calculated, and then a mask is applied. A train and test dataset for the feature vectors is created on which classification is performed, and an accuracy list is updated for the iteration. In the next iteration, the bandwidth is incremented by including the next consecutive band, the first pass of search. In the second pass, a similar search starts from the second band, and so on. This search is performed for all ($2^9-1=511$) feature masks, excluding the zero mask. This search outputs fewer combinations of bands with larger sizes than $b_w$. These bands are stacked together, and the iteration continues. During the second run with larger bandwidths, which is shown as "Coarse" in Fig.~\ref{fig_feature_overview} (a), the difference is that bands are not merged; that is, in the first pass, the features for the first and the second bands are first calculated separately and then concatenated together. The computation could explode as the process has to be performed for masked features of two or more bands. We only selected the best performers during the "Fine" bands search to counter this problem. In our case, the combination of two bands in such a way yielded a performance of above 95.0 \% in both soft fruits, which is our stopping criteria. The number of iterations for model training was set to 100 epochs, and the batch size was 32, which was found through hyperparameter optimization. The final output from the algorithm, in our case, is the set of bands with minimum or no overlaps, and together, their feature data produce classification accuracy above the required performance threshold.

\begin{figure}[htpb!]
    \centering
\begin{subfigure}[tb!]{\textwidth}
    \centering
        \includegraphics[scale=0.34]{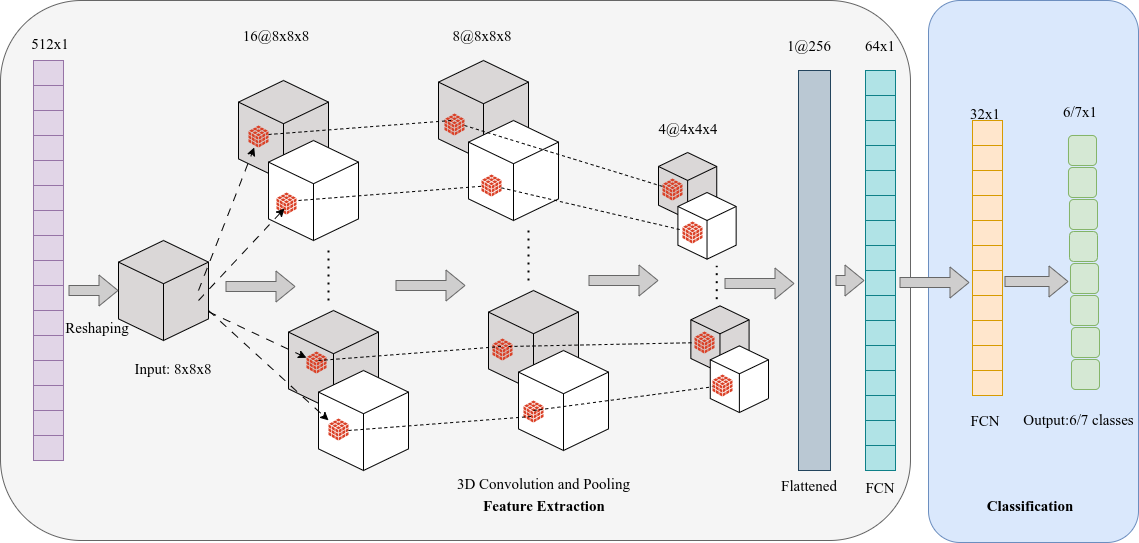}
        \caption{3D-CNN architecture.}
        \label{fig_3DCNN_arch}
\end{subfigure}

\begin{subfigure}[tb!]{\textwidth}
    \centering
    \includegraphics[scale=0.375]{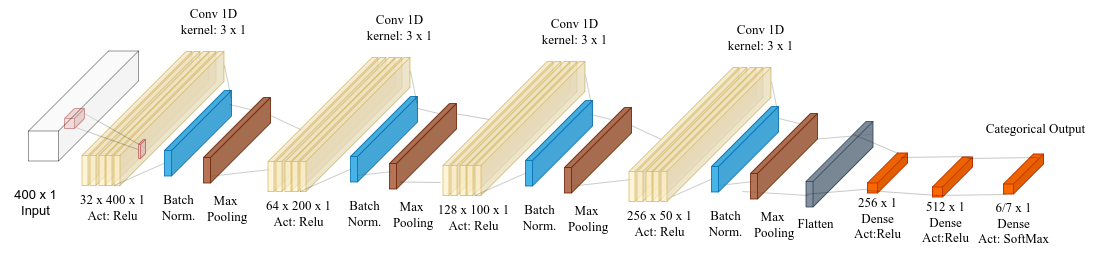}
    \caption{1D-CNN architecture.}
    \label{fig_1DCNN_arch}
\end{subfigure}
\caption{CNN architecture for classification of full-spectral volume pixels dataset.} 
\label{fig_cnn_archs}
\end{figure}

\subsubsection{Baseline Classifiers}
\label{sec_CNN_archs}

A comparative analysis of the proposed method is performed with CNN and SVM models. In this section, we describe two CNN models; both incorporate full-spectrum voxel data. The first model reshapes the input data into a 3D cube and then applies 3D convolution to the cube. This reshaping brings changes in the neighborhood such that distant data bands get closer enough to each other and lie inside the convolution kernel stride, making the potentially higher covariant bands more significant after convolution and pooling operations. Our images have a spectral resolution 400, which is extended to 512 by zero-padding after the last band to facilitate the data restructuring. The packing of length 512 hyperspectral vector into a ($8 \times 8 \times 8$) 3D Cube as shown in Fig.~\ref{fig_cnn_archs} (a). The model comprises configurable numbers of convolution (Conv3D) layers before the max pooling (MP3D) layer, after which it is flattened, and an FCN is applied with a softmax output layer. 

Similarly, a 1D-CNN architecture is also employed, whose architecture is shown in Fig.~\ref{fig_cnn_archs} (b). The model takes in the signature vectors and performs convolution through a set of convolutional layers; after the batch normalization, the max-pooling is applied, which is finally fed into the dense layers after passing through the flattening and dropout layers. The output of the model is in the categorical class format. However, the number of output classes for strawberry and tomato are different, which makes the output size different, as shown in Fig.~\ref{fig_cnn_archs}. The activation function in both 1D and 3D models is ReLU, except for the last layer, which has Softmax activation.

Finally, an SVM classifier with a polynomial kernel of degree 3 is also employed. Like the CNN models, the SVM classifier also takes the full-spectrum input data.

\subsubsection{Performance Metrics}
Accuracy scores are employed as performance metrics for the proposed feature extraction process, shown in Fig.~\ref{fig_feature_overview}. In addition to accuracy, which is sensitive to imbalance class distribution \cite{Petra2020}, we chose to express the performance through $\kappa$ coefficient \cite{Petra2020}; therefore, the model used in features extraction based on accuracy is later evaluated by $\kappa$. The accuracy is given as $Accuracy=\frac{(TP+TN)}{(TP+TN+FP+FN)}$, where TP is True Positive, FP is False Positive, FN is False Negative, and TN is True Negative. 

The $\kappa$ metric considers the probability of a chance agreement between the classifier and the ground truth chance: $\kappa$ measures the agreement between two evaluators, each dividing N items into mutually exclusive categories C. The definition of $\kappa$ is given as $\kappa=1-\frac{(1-p_o)}{(1-p_e)}$, where $p_o$ is the relative agreement of observations between rating systems, $p_e$ is the hypothetical probability of random agreement, and the observed data are used to calculate the probability of every observer observing each category randomly. If the rating is in complete agreement, then $\kappa=1$; otherwise, if it does not agree, then $\kappa=0$. These metrics measure the classification performance for strawberry and tomato datasets with GT annotations. 

\section{Results} \label{sec_results}
This section presents the classification results of strawberry and tomato maturity for various feature vector sets and bandwidths. The proposed method is compared with 1D-CNN, 3D-CNN, and SVM classifiers. 

\subsection{Feature Selection}

A preliminary analysis of the performance of individual features is investigated to ascertain the prominent ones through a combinatoric search on features, their masks, and reflectance subbands. Tab.~\ref{tab_straw_single_features} lists the features in the order of the classification model's test accuracy for strawberries. Argmax and maximum within the pigment band are the top two performers with 91.7 \% and 71.4 \% accuracy, respectively. Similarly, minimum and argmin are the best within the chlorophyll band, with an accuracy of 55.0 \% and 48.3 \%, respectively. The same feature performs poorly in the adjacent band ranges. The results show that features approximate pigment abundance change within the first range, while chlorophyll decomposition occurs in the second range.

\begin{table}[!htpb]
\centering
\scalebox{0.6}[0.6]{
\begin{tabular}{|c|c|c|c|c|c|}
\hline
\multicolumn{1}{|c|}{\textbf{Features}} &\textbf{Bands [nm]}  &  \textbf{Accuracy [\%]} &\textbf{Bands [nm])}  & \textbf{Accuracy [\%]}      \\ \hline
\multicolumn{1}{|c|}{argmax}     & 510--670 & \textbf{91.7} & 670--810 &  39.13   \\ \hline
\multicolumn{1}{|c|}{min}        & 530--670 & 27.5  & 670--790 & \textbf{55.0}  \\ \hline
\multicolumn{1}{|c|}{ max}        & 510--670 & 71.4  & 650--790 & 14.97  \\ \hline
\multicolumn{1}{|c|}{argmin}     & 530--670 & 69.5  & 670--790 & 48.30  \\ \hline
\multicolumn{1}{|c|}{kurtosis}   & 510--670 & 68.5  & 670--790 & 27.53  \\ \hline
\multicolumn{1}{|c|}{skewness}   & 510--670 & 64.7  & 670--790 & 21.25  \\ \hline
\multicolumn{1}{|c|}{mean}       & 510--670 & 59.9  & 670--810 & 30.91  \\ \hline
\multicolumn{1}{|c|}{median}     & 510--670 & 57.4  & 690--720 & 15.94  \\ \hline
\multicolumn{1}{|c|}{area}       & 510--670 & 39.1  & 670--790 & 30.91  \\ \hline
\end{tabular}}
\caption{Performance evaluation of individual feature within pigment window in strawberry dataset.}
\label{tab_straw_single_features}
\end{table}

The feature selection steps are similar for both strawberry and tomato datasets. The best individual feature's performance and corresponding band ranges are given in Tab.~\ref{tab_tom_individual_features}, ordered by accuracy. Similar to the strawberry dataset, there are two prominent subband regions; one ranges between 470 and 670 nm, and the other lies between 650 and 830 nm. The results of individual feature performance are given in Tab.~\ref{tab_tom_individual_features}. Argmax overperforms other features within the pigment band with an accuracy of 90.6 \%, while the 

\begin{table}[!htpb]
\centering
\scalebox{0.6}[0.6]{
\begin{tabular}{|c|c|c|c|c|c|}
\hline
\multicolumn{1}{|c|}{\textbf{Features}} &\textbf{Bands [nm]}  &  \textbf{Accuracy [\%]} &\textbf{Bands[nm]}  & \textbf{Accuracy [\%]}      \\ \hline
\multicolumn{1}{|c|}{argmax}     & 550--670 & \textbf{90.6} & 670--830 &  47.22   \\ \hline
\multicolumn{1}{|c|}{area}       & 490--550 & 73.33  & 660--690 & \textbf{58.33}  \\ \hline
\multicolumn{1}{|c|}{min}        & 510--650 & 77.22  & 650--790 & 55.55  \\ \hline
\multicolumn{1}{|c|}{ max}        & 470--530 & 74.45  & 650--790 & 30.00  \\ \hline
\multicolumn{1}{|c|}{kurtosis}   & 530--750 & 77.22  & 650--830 & 55.55  \\ \hline
\multicolumn{1}{|c|}{mean}       & 470--550 & 75.00  & 650--690 & 56.66  \\ \hline
\multicolumn{1}{|c|}{median}     & 570--550 & 74.44  & 670--750 & 56.66  \\ \hline
\multicolumn{1}{|c|}{skewness}   & 610--690 & 73.33  & 650--730 & 51.11  \\ \hline

\multicolumn{1}{|c|}{argmin}     & 550--610 & 62.77  & 650--790 & 21.66  \\ \hline

\end{tabular}}
\caption{Performance evaluation of individual features within pigment window in the tomato dataset.}
\label{tab_tom_individual_features}
\end{table}

\noindent area under the reflectance signature within the Chlorophyll band performs better than others with an accuracy of 58.3 \%. The variance in band ranges of the tomatoes is higher than that of the strawberries. The tomato features have a median accuracy of 75.0 \% compared to 64.7 \% in the pigment band, which implies better individual performance in tomatoes. Similarly, the median accuracy is 55.5 \% compared to strawberry's 30.9 \%.

A combination of features is explored; Tab.~\ref{tab_features_comb1} lists the combination of best-performing features and their corresponding bands by accuracy in strawberries. We observe the presence of argmax in all top combinations within the range of 490-670 nm. Maximum and argmax are combined to yield the best-performing 95.2 \% within its 510-670 nm subband. Similarly, within the 650--790 nm range, all listed combinations bear argmin and minimum. The maximum accuracy approaches 80.0 \% for the combination of minimum 
\begin{table}[!htpb]
\centering
\scalebox{0.49}[0.49]{
\begin{tabular}{|c|c|c|c|c|c|c|c|}
\hline
\multicolumn{1}{|c|}{\textbf{Features}} &\textbf{Bands [nm]}  & \textbf{Accuracy [\%]} & \textbf{Features} &\textbf{Bands [nm]}  & \textbf{Accuracy [\%]}      \\ \hline
\multicolumn{1}{|c|}{max,argmax}                     & 510--670 & \textbf{95.18} & min,argmin   & 670--790 & \textbf{79.71}    \\ \hline
\multicolumn{1}{|c|}{min, argmax, argmin, median, area} & 530--670 & 94.67 & max, min, argmax, argmin, mean, median & 670--790 & 79.22   \\ \hline
\multicolumn{1}{|c|}{min,argmax}                     & 490--670 & 94.61 & max, min, argmax, argmin & 670--790 & 78.74   \\ \hline
\end{tabular}}
\caption{Features, band range, and accuracy for variable band ranges within pigments window in the strawberry dataset.}
\label{tab_features_comb1}
\end{table}

\noindent and argmin; all other combinations constitute more features and yet underperform compared to it. Like strawberries, the combination of features in tomatoes contains argmax in top performing combination within the pigment band as shown in Tab.~\ref{tab_tom_features_comb1}, and the minimum is common in all top performers from the Chlorophyll band. Argmax and argmin are the best feature combinations in the 510--650 nm pigment subband, with an accuracy of 94.3 \%. The Chlorophyll band has a set of max, min, argmax, and argmin within the subband 650--770 nm, which achieves 72.2 \% accuracy. 

\begin{table}[!htpb]
\centering
\scalebox{0.515}[0.52]{
\begin{tabular}{|c|c|c|c|c|c|}
\hline
\multicolumn{1}{|c}{\textbf{Features}} &\textbf{Bands [nm]}  & \textbf{Accuracy [\%]} & \textbf{Features} &\textbf{Bands [nm]}  & \textbf{Accuracy [\%]}      \\ \hline
\multicolumn{1}{|c}{argmax,argmin}                     & 510--650 & \textbf{94.25} & max, min, argmax,argmin    & 650--770 & \textbf{72.22}    \\ \hline
\multicolumn{1}{|c}{argmax,argmin} & 450--670 & 91.55 & max, min, argmax, median, area, skewness & 650--790 & \textbf{72.22}   \\ \hline
\multicolumn{1}{|c}{argmax, argmin, area}                     & 510--670 & 90.99 & max, min, argmax, mean, kurtosis & 650--830 & \textbf{72.22}   \\ \hline
\end{tabular}}
\caption{combination of features and their band range, and accuracy in chlorophyll band in the tomato dataset. }
\label{tab_tom_features_comb1}
\end{table}

The next level of feature combination for the strawberry classification yields results listed in Tables \ref{tab_straw_single_features} and \ref{tab_features_comb1}. It shows higher dependence on maximum and argmax in the pigment band and various features within the chlorophyll band. The best accuracy of 98.7 \% is achieved by combining the maximum and argmax selected from pigment and minimum, maximum, argmax, argmin, mean, median, and area under the curve from the chlorophyll band. The second best-performing combination approaches 98.0 \% accuracy, utilizing only two features as above from pigment bands, combined with the minimum and argmin from the chlorophyll band. Due to the minimum number of features and the accuracy of approaching the best one, we select the feature set given in the second row of Tab.~\ref{tab_features_comb1} for classification, as it would be cost-effective to build and deploy.

 \begin{table}[!htpb]
\centering
\scalebox{0.6}[0.6]{
\begin{tabular}{|c|c|c|c|c|c|}
\hline
\multicolumn{1}{|c|}{\textbf{Features}} &\textbf{Bands [nm]} & \textbf{Features} &\textbf{Bands [nm]}  & \textbf{Accuracy [\%]}      \\ \hline
\multicolumn{1}{|c|}{max, argmax} & 510--670 & max, min, argmax, argmin, mean, median, area & 670--790 & \textbf{98.73}    \\ \hline
\multicolumn{1}{|c|}{\textbf{max, argmax}} & 510--670 &  \textbf{min, argmin} & 670–790 & 98.04    \\ \hline
\multicolumn{1}{|c|}{max, argmax} & 510--670 & max, min, argmax, argmin   & 670--790 &     97.58\\ \hline
\end{tabular}}
\caption{Features, bandwidth, and accuracy for best-performing subbands within chlorophyll window in the strawberry dataset. The selected features and highest accuracy are shown in boldface. }
\label{tab_feature_comb_straw_results}
\end{table}

The maximum accuracy of 96.8 \% is achieved in tomatoes with the argmax and argmin selected from the pigment bands and combined with the maximum, minimum, argmax, and argmin from the chlorophyll band as given in Tab.~\ref{tab_tom_features_comb1}. The second combination with the highest accuracy is close to 96.5 \% by cumulative features given in the first row of Tab.~\ref{tab_tom_features_comb1}.

\begin{table}[!htpb]
\centering
\scalebox{0.6}[0.6]{
\begin{tabular}{|c|c|c|c|c|c|}
\hline
\multicolumn{1}{|c|}{\textbf{Features}} &\textbf{Band Range [nm])} & \textbf{Features} &\textbf{Band Range [nm]} & \textbf{Accuracy [\%]}      \\ \hline
\multicolumn{1}{|c|}{\textbf{argmax,argmin}} & 510--650 &  \textbf{max,min,argmax, argmin} & 650-770 & \textbf{96.81}    \\ \hline
\multicolumn{1}{|c|}{argmax,argmin} & 510--650 & max, min, argmin & 650-770 & 96.54    \\ \hline
\multicolumn{1}{|c|}{argmax,argmin} & 510--650 & max, min, argmax, mean   & 650--790 &     95.14\\ \hline
\end{tabular}}
\caption{Features, band range, and accuracy for variable bandwidth in chlorophyll band in the strawberry dataset. The selected features and highest accuracy are shown in boldface.}
\label{tab_tom_combined1}
\end{table}

\begin{figure}[htpb!]
    \centering
\begin{subfigure}[tb!]{0.49\textwidth}
    \includegraphics[scale=0.4]{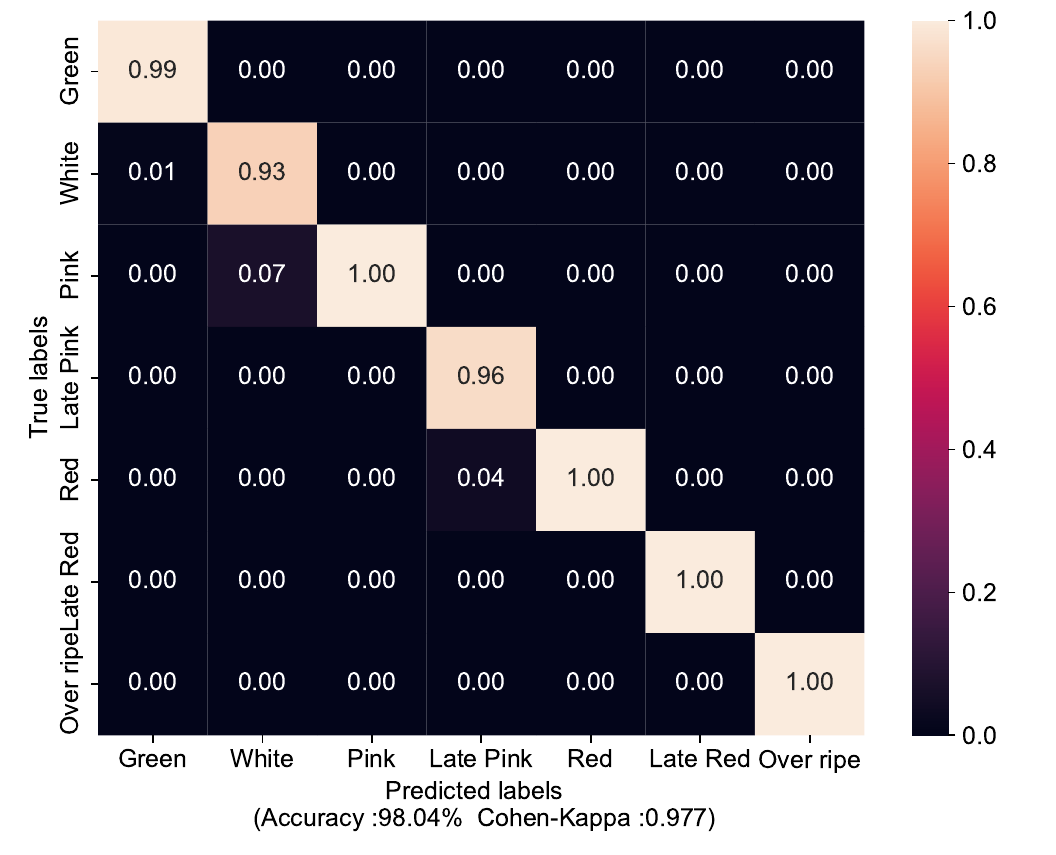}
\end{subfigure}
\begin{subfigure}[tb!]{0.49\textwidth}
    \includegraphics[scale=0.4]{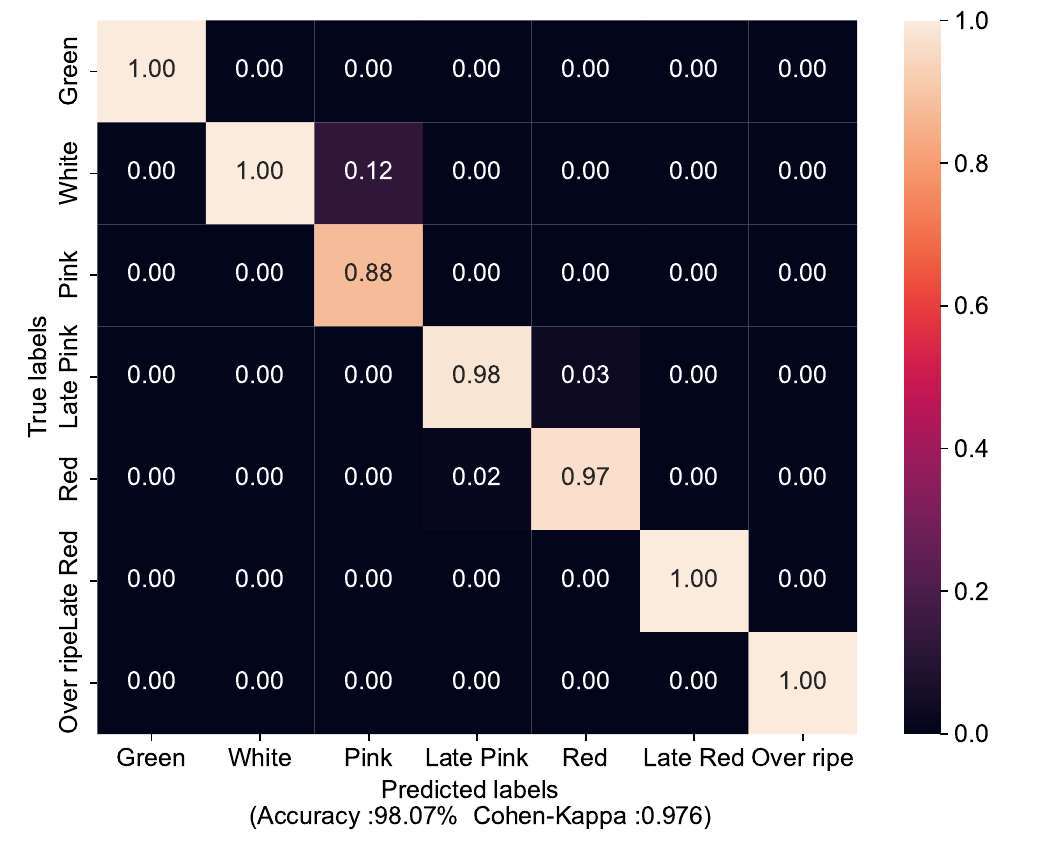}
\end{subfigure}
\caption{Heat plots for FCN (left) and SVM (right) predicted results for the selected features of strawberries.}
\label{fig_straw_heatmap}
\end{figure}

\begin{figure}[htpb!]
    \centering
\begin{subfigure}[tb!]{0.49\textwidth}
    \includegraphics[scale=0.4]{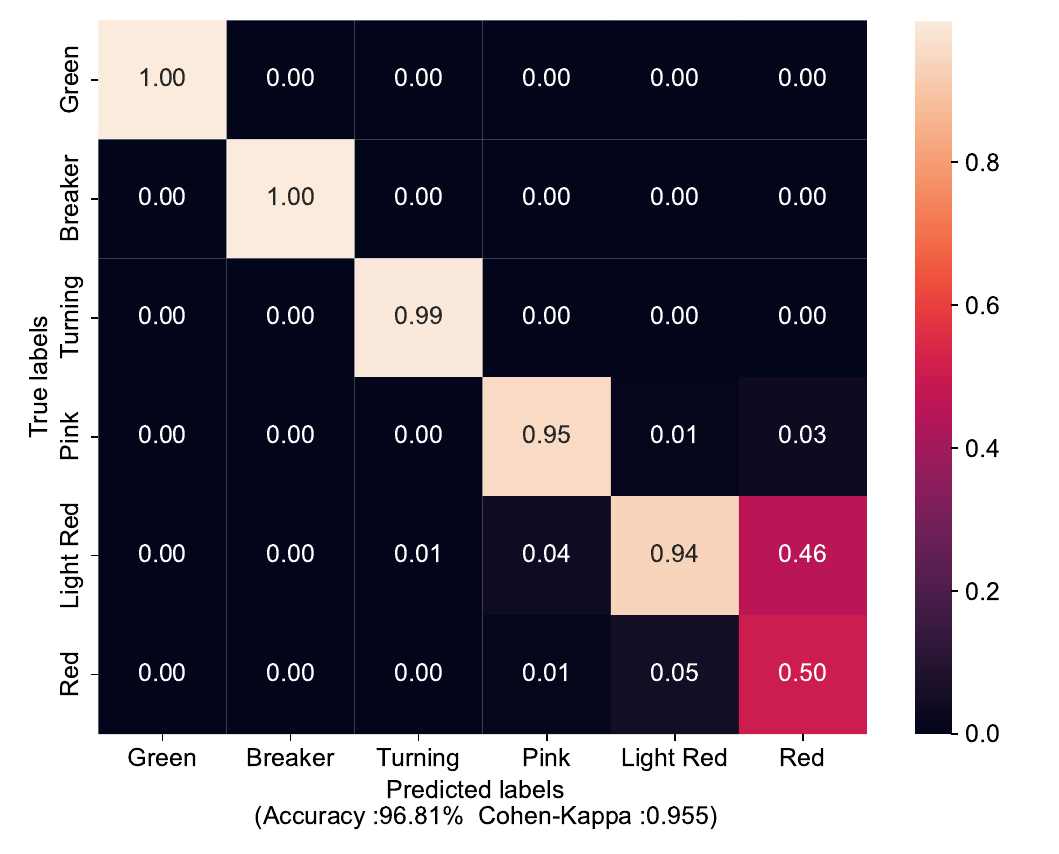}
\end{subfigure}
\begin{subfigure}[tb!]{0.49\textwidth}
    \includegraphics[scale=0.4]{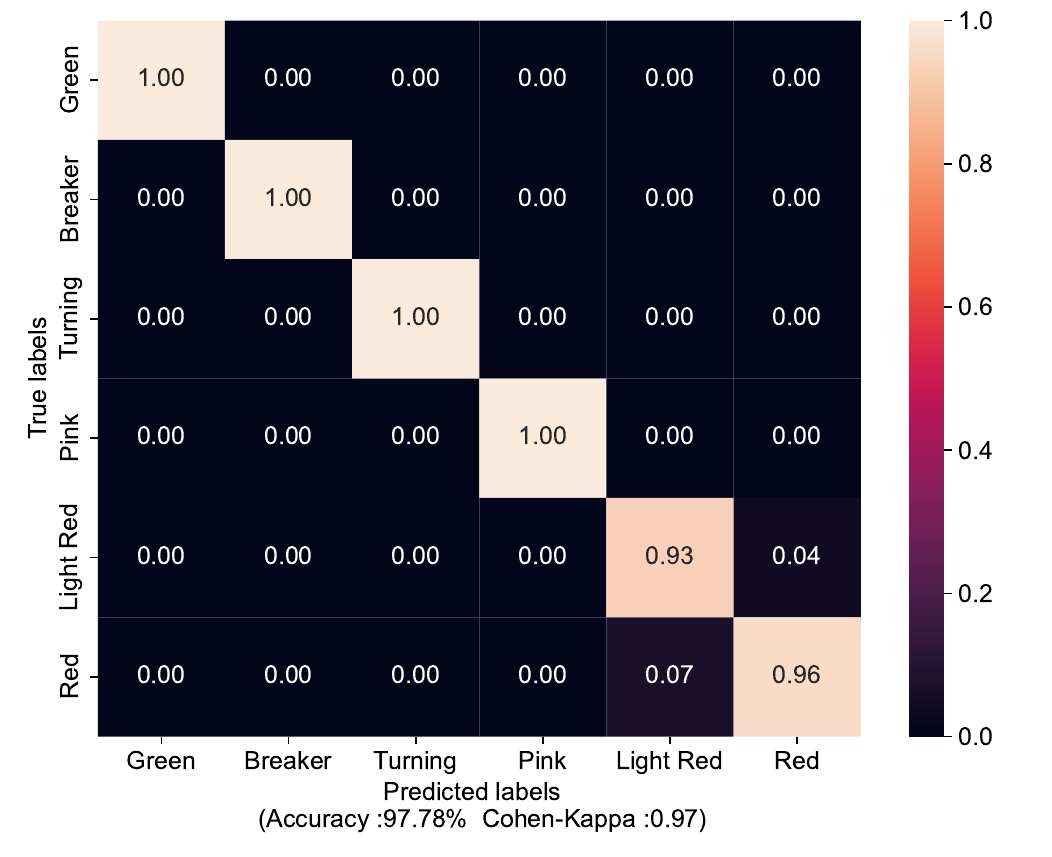}
\end{subfigure}
\caption{Heat plots for FCN (left) and SVM (right) predicted results for the selected features of tomatoes.}
\label{fig_tom_heatplots}
\end{figure}

Fig.~\ref{fig_straw_heatmap} and \ref{fig_tom_heatplots} show the classification results of both fruits in the heat plot for their respective selected features in Tab.~\ref{tab_feature_comb_straw_results} and \ref{tab_tom_combined1}. In strawberries, the chosen features, such as max and argmax in the pigment band, are constant for the top three results, whereas in tomatoes, the similar common features are argmax and argmin. However, the Chlorophyll bands have more than two features in strawberries and tomatoes. Later, the features for both bands are concatenated together to create a single vector in all cases. The best maturity classification has an accuracy of 98.7 \% in strawberries. The nominal error appears due to confusion between \emph{Late-Red} and \emph{Overripe}, \emph{Red} and \emph{Late-Red}, \emph{Pink} and \emph{Late-Pink}, and \emph{Green} and \emph{White} classes. \emph{White} and \emph{Red} strawberries are classified ideally in this feature vector. The second row of Tab.~\ref{tab_feature_comb_straw_results} lists the next feature vector, which merges pigment features with min, argmin from the Chlorophyll bands and yields an accuracy of 98.0 \%. Furthermore, these four attributes are related to two single wavelengths in given bands compared to more aggregated features requiring complete hyperspectral data for computing mean and median. Our selection criteria for the model are based

 \noindent on higher accuracy and the least number of features. Therefore, we select the feature set in the second row of Tab.~\ref{tab_feature_comb_straw_results}. The same feature set is classified with an SVM classifier. It achieved an accuracy of 98.0 \%. This is shown in Fig.~\ref{fig_straw_heatmap} (b). The last row comprises max, argmax max, min, argmax, and argmin in the Chlorophyll band and has an accuracy of 97.5 \%. Therefore, it is not considered further.

Similarly, for tomatoes, the classification results for the selected characteristics in Tab.~\ref{tab_tom_combined1} are shown in Fig.~\ref{fig_tom_heatplots}. The values of the pigment band features, such as argmax and argmin, are the same for the three top-performing results. Similarly, the chlorophyll band has min and max common in the top three. Fig.~\ref{fig_tom_heatplots} (a) shows the feature vector results, including the chlorophyll band's maximum, minimum, and argmin. The dual-band features are concatenated to create a single vector in all cases. The classification is 96.8 \% accurate. Nominal errors appear due to classification errors between the \emph{Pink}, \emph{Light-Red}, and \emph{Red} classes. The poor performance is in the classification of \emph{Red} tomatoes, where 46.0 \% cases are classified as \emph{Light-Red} and 3.0 \% as \emph{Pink}. \emph{Green}, \emph{Breaker}, and \emph{Turning} have accuracy between 99.0 \% and 100.0 \%. Fig.~\ref{fig_tom_heatplots} (b) shows results of the same selected feature, but the classification model used is SVM instead of FCN. 
It achieved an accuracy of 97.7 \% as shown in Fig.~\ref{fig_tom_heatplots} (b). Therefore, FCN and SVM perform equally well on selected features, with SVM having a slight edge over the FCN classifier. The second row in Tab.~\ref{tab_tom_combined1} shows the next characteristic vector, which merges pigment 

\begin{figure}[htpb!]
    \centering
    \includegraphics[scale=0.285]{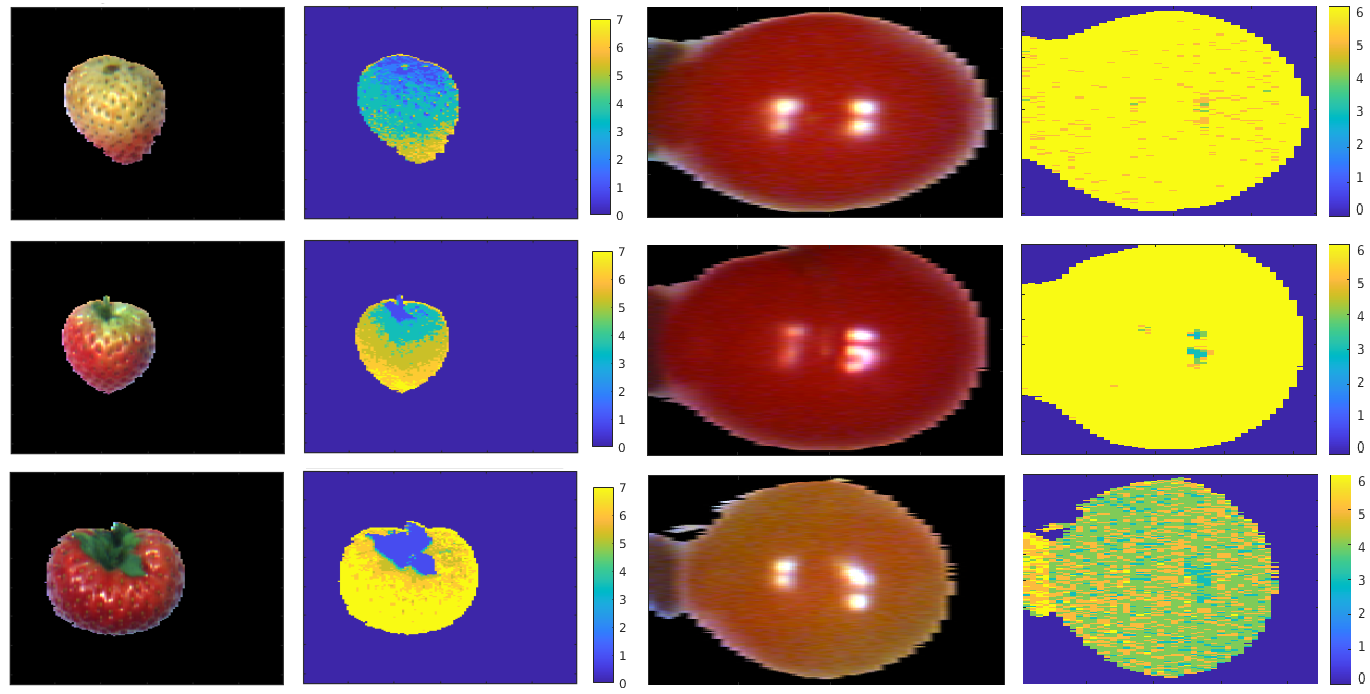}
    \caption{Prediction of pixel-wise strawberry and tomato maturity classes are shown in seven and six classes, respectively, with zero being the background.} 
\label{fig_straw_tom_pixel}
\end{figure}

\noindent characteristics with minimum argmin from the chlorophyll band and has 96.5 \% accuracy. The third row in Tab.~\ref{tab_tom_combined1} shows higher classification errors in distinguishing between the last three classes. The characteristic vector includes maximum, minimum, argmax, and mean in the chlorophyll band, and its accuracy is 95.1 \%. 

The selected features for both fruits were employed to predict background-subtracted hyperspectral images, as depicted in Fig.~\ref{fig_straw_tom_pixel}. It is visually demonstrated that pixel-level classification is plausible for future investigations. The class map ranges from one to six or seven ripeness levels, depending on the fruit, with zero being the background.

\subsection{Baseline Methods}

A comparison of the proposed method with 1D and 3D CNN models and SVM is included in this section. The CNN models used for classification are custom-built, their architectures are covered in Section \ref{sec_CNN_archs}. A summary of results according to our performance metric is shown in Tab.~\ref{tab_models_compare}. The prediction FPS for images in our dataset is also included. An estimate of FPS is also computed for Real-time HSI (\textbf{RT-HSI}), i.e., Ultris S5 resolution ($290\times275\times51$), which has significantly lower spectral resolution than our camera.

\begin{table}[!htpb]
	\centering
\scalebox{0.515}[0.515]{
\begin{tabular}{|l|c|c|c|c|c|c|c|c|}
\hline
\multicolumn{1}{|c|}{Classification Method} & \multicolumn{4}{c|}{Strawberry} & \multicolumn{4}{c|}{Tomato}\\
\cline{2-9}
\multicolumn{1}{|c|}{ } & Accuracy [\%] & $\kappa$ & Pred. FPS & RT-HSI Pred. FPS & Accuracy [\%] & $\kappa$ & Pred. FPS & RT-HSI Pred. FPS\\
\hline
HSI SVM (Full-spectrum)   & 89.37   & 0.867 & 1.16 & 3.94 & 91.67   & 0.896 & 1.39 & 4.72 \\ \hline
HSI 1D CNN (Full-spectrum)& 87.92     & 0.847 & 1.14 & 3.87 & 88.33   & 0.852 & 1.29 & 4.36 \\ \hline
HSI 3D CNN (Full-spectrum) & 89.86 & 0.877 & 0.26 & 0.90 & 90.34   & 0.878 & 0.32 & 1.07  \\ \hline
Proposed Method (FCN)  & 98.04 & \textbf{0.977}& 0.65 & 2.19 & 96.81   & 0.955 & 0.70 & 2.37\\ \hline

Proposed Method (SVM)  & \textbf{98.07} & 0.976 & \textbf{12.88} & \textbf{43.71} & \textbf{97.78}   & \textbf{0.970} & \textbf{12.78} & \textbf{43.37}  \\ \hline

\end{tabular}}
\caption{Classification results of strawberry and tomato in terms of accuracy and $\kappa$ for different classification methods are listed. The model prediction (Pred.) FPS on images from our dataset is also given for every method. An estimate of prediction FPS on an RT-HSI camera, i.e., Ultris S5, is also listed, as per its resolution, to highlight that the proposed method could be suitable for real-time applications.}
\label{tab_models_compare}
\end{table}

The 1D-CNN has a classification accuracy of 89.4 \% in strawberries. There is more considerable confusion between \emph{Pink} and \emph{White}, \emph{White} and \emph{Green}, \emph{Red} and \emph{Late-Red}, and \emph{Overripe} and \emph{Late-Red} classes, respectively. The best performance is in the \emph{Green} and  \emph{Late-Pink} classes, respectively. A slight slump in the performance is observed in tomato maturity classification; the overall accuracy drops to 88.3 \%. The significant classification error is between the \emph{Turning} and \emph{Breaker} and \emph{Pink} and \emph{Turning} classes. The 3D-CNN model has improved strawberry classification performance with an accuracy of 89.8 \% compared to the 1D counterpart. The confusion classes are similar to 1D-CNN except for the slightest confusion in the \emph{Late-Red} class. The improvement in the tomatoes is also observed here; for example, accuracy increased to 90.3 \%.
Contrary to 1D-CNN results, the worst score is in the \emph{Overripe} class, which performs decently in the 1D case. Although 3D-CNN overperformed its 1D counterpart in strawberries and tomatoes, its accuracy difference from SVM's is slim. The accuracy in SVM is 89.3 \%. The SVM performs worst in classifying \emph{White} and \emph{Pink} classes but performs well in ripe classes, i.e., \emph{Red} and above. SVM performs the best among the SOTA methods in tomatoes with an accuracy of 91.6 \%. The \emph{Overripe} class has the highest error in SVM, followed by the \emph{White} one. The proposed method outperforms all these methods by a significant margin of accuracy between 9.0--11.0 \% in strawberries and 5.0--6.0 \% in tomatoes. Although the classification error of the proposed method is marginal in strawberries, it has worse performance in differentiating between \emph{Red} and \emph{Light-Red} classes for tomatoes to the extent that the accuracy of the \emph{Red} class is 53.0 \%, as shown in Fig.~\ref{fig_tom_heatplots}. According to the trend in results of CNN and SVM classifiers for strawberry and tomato maturity classification, the highest errors are in the \emph{White}, \emph{Pink}, and \emph{Overripe} classes. However, SVM error for strawberries is distributed across \emph{White}, \emph{Pink}, \emph{Late-Pink}, and \emph{Red} classes in strawberries and concentrates only at the \emph{Breaker} class in tomatoes.

\section{Discussion}
This paper empirically demonstrates that a subset of features from hyperspectral images is sufficient to establish a maturity classification system for strawberries and tomatoes. A comprehensive feature extraction algorithm was developed, which seeks the best combination of variable bandwidths and predefined feature vectors associated with them, such as reflectance values at the peak or trough, their corresponding wavelength values and other statistical measures aggregated across the subband reflectance signature. It is shown that the wavelength position of extremums in different bands is vital information for the maturity classification problem. Moreover, the reflectance intensity at these points is essential supplementary information that improves the model accuracy when combined with extremum position data.

We employed several statistical features for our analysis but observed that higher-order measures such as skewness and kurtosis underperformed. The max, min, argmax, and argmin. Luckily, the prominent features are related to a single wavelength, making the feature selection more straightforward. It should be noted that we found two wavelengths, but they vary from pixel to pixel. Therefore, hyperspectral data is required. A comparative analysis was performed with CNN models and SVM with full-spectrum reflectance data. 1D and 3D CNN models are constructed for comparative analysis, which takes the full-spectrum data as input for training and testing the model. Despite rich spectral input to these models, the features extracted in the proposed method outperform them significantly in both fruits. The analysis of results manifests that strawberries have higher confusion between \emph{White} and \emph{Pink} classes in all classification models except the proposed method. The increased classification errors appear in tomato for \emph{Breaker} class in all models except the proposed model, which performs poorly in the \emph{Red} class. 
Eventually, the results show that the selected feature of our method will perform equally well with any non-linear classifier. Despite having the same level of accuracy, the SVM with a polynomial kernel has higher computational efficiency than the FCN model employed in our proposed method. 

Hyperspectral imaging typically entails a relatively lower image acquisition speed; therefore, the number of images is fewer than conventional color images. The proposed method requires a relatively smaller amount of data for classification compared to SOTA DL methods. The feature selection for given fruits requires less preprocessing, which is typically necessary for full-spectrum reflectance input. This is evident in the time performance comparison of the proposed method with the SOTA, given in prediction FPS, and its maximum speedup is more significant than eleven compared to its counterparts. Constructing a real-time and cost-effective solution for maturity classification using the proposed method is plausible.

\section{Conclusions and future work}

We developed a search-based feature extraction method for strawberry and tomato maturity classification. A fixed number of masked features is calculated for combinatorically varying bandwidths, and their classification results are recorded in each iteration. The search was performed in full-spectrum and combinations of sub-spectrums. Unlike conventional band selection algorithms, we demonstrate that extremum points position data in some subbands provided significantly correlated information with maturity stages. The best-performing features and associated bandwidths were selected for both fruits. A comparative analysis was performed with CNN models and SVM. The proposed method outperformed the other classifiers with accuracy in strawberries up to 98.04 \% compared to the 1D-CNN model's 87.92 \% and 89.86 \% of the 3D-CNN classifier's accuracy. In tomatoes, it has 97.78 \% compared to the 1D-CNN model's 88.33 \% and 90.34 \% of the 3D-CNN classifier's accuracy. The SVM classifier achieved 89.37 \% and 91.67 \% accuracy in strawberries and tomatoes, respectively. It is observed that the classes where the sample size is smaller possess lower accuracy.

The FCN model used for the feature selection process yields accuracy similar to the SVM, i.e., 98.07 \% and 97.78 \% for strawberries and tomatoes, respectively. This manifests that the features are distinctive and sufficient enough for good classification by any non-linear model. The computation efficiency of all classifiers was investigated. The proposed method with an SVM classifier was the best performer, having approximately 13 FPS compared to the next best 1.16 from full-spectrum SVM. 

All baseline methods required complete hyperspectral voxel reflectance data as input. The trained model was then applied to background removed images to produce a ripeness map for both fruits and calculate the pixel-based ripeness for classification visualization. Eventually, the proposed method reduces the preprocessing requirement and ensures the simplicity of the prediction model, which will ease the deployment process. This would enable us to construct a high-speed, cost-effective solution for maturity classification problems such as selective harvesting and QC in packaging sites. 

An interesting future work is to develop a regression model to estimate these peak and trough points from multispectral data. This would make the hardware cost-effective and more acceptable for commercial applications. The number of images employed for one type of fruit was around 600. Therefore, the DL models underperformed compared to the proposed method, so another future investigation would roughly quantify the number of images required for the DL model to compete closer to the proposed method.

\section*{Acknowledgements}

This work is partly supported by Innovate UK grant 10057282 funding and Research England Expanding Excellence in England for Lincoln Agri-Robotics (LAR). We want to acknowledge the contributions of the University of Lincoln staff, including Sophie Bowers, who curated fruit samples, and Dr. Robert Lloyd and Andrew Ham, who provided support related to laboratory equipment.

 \bibliographystyle{elsarticle-num} 
 \bibliography{references}
\end{document}